**TECHNICAL NOTE**

# Tabular Foundation Models in Financial Predictions

Ed deHaan
*Stanford University*
eddehaan@stanford.edu

Allen H. Huang
*Hong Kong University of Science and Technology*
allen.huang@ust.hk

November 10, 2026

**Abstract**

This technical note introduces tabular foundation models (TFMs) to accounting research and evaluates their performance in one common benchmark task: predicting firms' next-year earnings changes. TFMs produce numeric predictions and learn in context, without post-release training or tuning, making them natural and easy to use for forecasting. The TFMs we study are pretrained on synthetic tabular data, so they avoid look-ahead bias from memorizing firms' outcomes. We compare Google Research's TabFM with a fine-tuned open-weight large language model developed in recent accounting research and with several tree-based models. TabFM is more accurate than all comparison models. It is also faster and less computationally intensive than the large language model. Several tree-based specifications produce predictions more quickly than TabFM, so choosing between them involves a trade-off between accuracy and speed. We conclude that TFMs have promise as a new tool for forecasting in accounting and finance. However, our evaluation is based on a single benchmark task, and no class of models is likely to dominate in all settings.

----

We appreciate helpful comments from Weihao Kong and Abhimanyu Das of Google Research, Chanseok Lee, Bradford Levy, and Yi Yang. We thank Yiwei Dou and Qishen Fu for very kindly sharing their data and insights, and Wenbo Wang for excellent research assistance.

## 1. Introduction

Much of empirical accounting research uses data organized into tables, with firm-period observations in rows and financial characteristics in columns. Researchers use these data to forecast earnings, assess financial distress, and predict other economic outcomes. Tabular foundation models (TFMs) offer a new approach to these applications. They are designed for structured tabular data and are pretrained across hundreds of millions of synthetic datasets to generate numeric predictions. This technical note introduces TFMs, explains their relationship to conventional machine learning and large language models (LLMs), and illustrates their application to earnings prediction. We implement Google Research's TabFM and compare its predictive performance and costs with those of models examined in recent research: a fine-tuned open-weight large language model (Dou & Fu 2026, hereafter "DF") and several tree-based models (Chen et al., 2022; Levy 2026). The note does not aim to establish that any one forecasting technology dominates another; such a conclusion cannot be drawn from one study, especially given how rapidly technologies are evolving. Rather, our goal is to gauge whether and when TFMs can provide a useful addition to an earnings forecaster's toolbox.

TFMs generate predictions through in-context learning (ICL), which is taught to the model before its public release. In this "pretraining" stage, the developer trains the model on many prediction tasks so that its parameters encode a general procedure for extracting predictive relationships from a new dataset. In essence, the model learns how to learn from data, an approach known as meta-learning (Hospedales et al., 2022; Müller et al., 2022). Once the model is released, a user can apply it to a new prediction problem without training or tuning it for that task. The user instead supplies two sets of observations: historical examples with known outcomes, and the observations to be predicted. The model applies its pretrained procedure to extract patterns from the historical examples and use them to generate the predictions. Conventional machine learning models, by contrast, have no pretraining so must be trained by the user on each new dataset, usually with a search over hyperparameter configurations. Open-weight large language models are pretrained, but fine-tuning for a specific task requires a user to adjust its billions of parameters.

TabFM's developers report that its pretraining datasets were dynamically generated by computer programs rather than collected from actual firms (Kong & Das 2026). Two implications follow. First, TabFM has no built-in accounting knowledge, so it does not know or understand what any column represents.[1] Like conventional machine learning models, whatever TabFM learns about how financial statements relate to future earnings must instead come from the numbers in the in-context training sample alone. Second, TabFM cannot have memorized firms' outcomes, so look-ahead bias from seeing outcomes can arise only from the in-context training sample, which the user controls. TabFM can therefore be more safely back-tested over historical periods in which a large language model could recall firms' outcomes from pretraining.[2]

We evaluate models on both predictive performance and cost. To compare with recent research, we assess models' ability to predict the direction of one-year-ahead earnings changes and measure performance using the area under the receiver operating characteristic curve (AUC).

[1] In practice, TabFM ignores column names, as do conventional machine learning models.

[2] Synthetic pretraining does not make TabFM immune to all forms of look-ahead bias. If the synthetic data resemble firms' actual earnings-generating processes and those processes change over time, in-sample tests could overstate the performance a forecaster could have achieved in real time. A related concern applies to most forecasting models, including the tree-based models, whose implementations and settings reflect users' choices based on current knowledge. A second issue is that TabFM did not exist at our forecast dates, which can be considered a form of technological look-ahead bias (Levy 2026). Finally, as with any model built by a third party, users must rely on the developers' account of how TabFM was pretrained. The developers report that pretraining used only synthetic tables (Kong & Das 2026; Kong et al. 2026). We have no reason to doubt these reports or the assurances the developers gave us in conversation (personal communication, October 6, 2026), and our tests below show no sign of memorization.

AUC captures how well a model ranks firm-years with earnings increases above those without.[3] We measure cost as each model's computing time and memory requirements, taking the pretraining of TFMs and large language models as given.

Our main analyses compare TabFM with five models. The first is the fine-tuned Llama model developed in DF (hereafter "Llama"). The other four are tree-based: DF's gradient-boosted decision tree ("GBDT-DF"); random forest ("RF") and stochastic gradient boosting ("SGB"), following Chen et al. (2022); and our re-estimation of GBDT using XGBoost ("GBDT-X"). To ensure comparability, we evaluate Llama and GBDT-DF using observation-level predictions that DF kindly provided. We generate our own predictions for RF, SGB, and GBDT-X, which allows us to use them in additional analyses.

Our analyses use the sample and 130 financial statement predictors provided by DF. The sample contains 62,796 firm-years with financial statements from 2005 through 2024, all from firms with December 31 fiscal year-ends. Our main tests use the 6,569 firm-years with 2023 and 2024 financial statements to predict earnings changes in 2024 and 2025, the years DF evaluate to avoid look-ahead bias from Llama's pretraining.

In our main tests, TabFM outperforms all comparison models, with a pooled AUC of 69.53% across 2023 and 2024. Llama follows at 67.88%, and the tree-based models range from 62.75% to 63.78%. TabFM's advantages of 1.66 percentage points over Llama and 5.75 to 6.78 points over the tree-based models are all statistically significant, as are Llama's advantages over the tree-based models. Differences among the tree-based models are generally insignificant. Each test year separately yields the same ordering of models.

[3] Predicting one-year-ahead earnings changes may not be the most relevant task in practice, and AUCs may not capture the practical value of differences in prediction performance. We use these methods for comparability with prior research. (e.g., Freeman et al. 1982; Ou & Penman 1989; Chen et al., 2022; DF; Levy 2026).

We evaluate costs based on processing time, and our comparisons across models are more nuanced. Producing the 2023 and 2024 forecasts takes TabFM about 12 minutes on one GPU, with 24.3 GiB of peak GPU memory.[4] SGB takes 7.5 minutes, while RF and GBDT-X take 130 and 140 minutes, respectively, and all three require less memory than TabFM. GBDT-DF takes longer, reflecting its deliberately broad configuration search. DF report that its configuration search takes about 120 hours on a 128-core CPU server, with about 100 configurations running in parallel. According to DF, fine-tuning Llama takes about 108 hours on an NVIDIA H200 GPU, and producing its forecasts takes about six minutes.

The models also incur their costs at different stages. The tree-based models spend almost all of their time on up-front training, after which incremental forecasts from the trained model are nearly instant. Llama's fine-tuning, which DF repeat for each test year, is likewise an up-front cost, and its forecasts then take a few minutes. TabFM has no training step, so its time recurs each time it produces a set of forecasts.

Weighing performance against cost, TabFM compares favorably with Llama on both, whereas the choice between TabFM and the tree-based models depends on how the forecasts will be used. Against the tree-based models, TabFM retains the highest AUC, but SGB has a shorter total time, and all of the tree-based models produce incremental forecasts almost instantly once trained. The tree-based models therefore have an advantage when a single trained model is used to produce many forecasts, whereas TabFM has an advantage when accuracy is the priority or when the training sample is updated frequently, because each update requires the tree-based models to repeat their training. Among the tree-based models, RF, SGB, and GBDT-X perform

[4] GPU denotes graphics processing unit; CPU denotes central processing unit, the computer's general-purpose processor; and RAM denotes random-access memory. Memory is measured in gibibytes (GiB). Peak GPU memory is the maximum memory in use on the GPU during a run, as recorded by nvidia-smi, less the memory already in use when the run starts.

similarly to GBDT-DF at lower cost.

The remainder of the note examines performance and practical choices in applying the models. Most of these analyses compare TabFM with RF, SGB, and GBDT-X. We exclude Llama and GBDT-DF because each analysis would require repeating costly fine-tuning or training, and because Llama cannot be evaluated before 2023 without potential exposure to memorization-induced look-ahead bias.

- **Variability over time.** Annual tests for 2011 through 2024 find that TabFM has a higher AUC than the tree-based models in every year, except for a slightly lower AUC than SGB in 2018. TabFM also has the smallest standard deviation of annual AUCs, so its higher average performance does not come with greater variability. All models' AUCs vary considerably across years, although all are well above random chance. The variation in performance across years suggests that results from only a few test years, including our main tests, should be interpreted with some caution.
- **Sensitivity to omitted predictors.** All four models are resilient to omitting potentially important predictors. We retrain the tree-based models and rerun TabFM after dropping the five predictors that DF identify as most important for Llama, and none has a significant decline in AUC. Keeping only the current and prior-year income statements (52 predictors, omitting all balance sheet variables) leaves the AUCs of TabFM and RF statistically unchanged, and it produces small but significant declines for SGB and GBDT-X. The largest declines come from dropping the 20 predictors that DF rank as most important, but even then no model loses more than 2.5 percentage points, and TabFM loses the least, at 1.3 points. TabFM has the highest AUC in every specification.
- **Smaller training samples.** All models also retain considerable predictive ability with

smaller training samples. AUCs decline as we reduce the training sample from the full sample to 5% of observations, but even at 5%, which is about 2,800 firm-years per test year, AUCs range from 59.5% to 64.5%. TabFM has the highest AUC at every sample size, and its 64.5% AUC in the 5% subsample exceeds the full-sample AUCs of all the tree-based models. TabFM's running time falls by 89%, to 1.3 minutes in the 5% subsample, and it is the fastest model at 60% of the sample and below. Tests that draw random subsets of firms yield similar inferences, even with as few as 100 firms.

- **Digit perturbations.** TabFM is not pretrained on actual firm outcomes, so it cannot have memorized them. Moreover, our annual tests find that TabFM's annual variability is roughly 90% correlated with tree-based models' annual variability, which is unlikely to be the case if TabFM's predictions stem from memorized outcomes. Still, for completeness we apply the least-significant-digit perturbation tests of Levy (2026), which are designed to detect memorization. Randomizing the last digit of each predictor lowers TabFM's AUC by only 0.38 percentage points, within the range of the tree-based models, whereas in Levy's tests the same perturbation removes all of GPT-4's predictive ability. Perturbing three and five digits removes more information and lowers all models' AUCs, but TabFM retains the highest AUC at every level. The tests therefore find no evidence of memorization.
- **Alternative tabular foundation models.** We compare our main TabFM specification with two other TFMs, TabICLv2 and TabPFN-3.5, and with alternative TabFM configurations, including the more recent TabFM 1.1.0. Because TFMs are at an early stage of development, we hesitate to draw strong inferences from these comparisons. All specifications have AUCs within two percentage points of one another, and all significantly outperform every tree-based model. Our main specification of TabFM has the highest AUC. TabICLv2 has the shortest running time, at 1.2 minutes, and TabPFN-3.5 has the lowest GPU memory use, at 3.8 GiB, while coming within one percentage point of TabFM's AUC. Different TabFM 1.0.0 specifications significantly outperform Llama, whereas TabICLv2, TabPFN-3.5, and the TabFM 1.1.0 specifications are statistically tied with it.

Overall, our evidence indicates that TFMs offer a promising tool for financial predictions with tabular data, despite not having any built-in accounting knowledge. In our benchmark task, TabFM is more accurate than one specification of an open-weight LLM and several tree-based models, and it achieves this without a configuration search. It retains its advantage with small samples and with relatively few predictors. TFMs can also be used to back-test forecast performance in historical data without concern that the model knows firms' outcomes from pretraining. The costs of TFMs and tree-based models arise at different stages: TabFM spends its time producing forecasts, whereas the tree-based models spend theirs on training, so the preferred model depends on how the forecasts will be used.

Our inferences are subject to several important qualifications. First, as DF note (page 10), no single benchmarking exercise can establish the superiority of one prediction technology over another, and it is unlikely that any single technology can dominate in all tasks, samples, periods, or configurations.[5] Relatedly, AI technologies are developing quickly, and it is quite possible that a newer open-weight model available today, fine-tuned as in DF, could outperform TabFM. Second, our comparisons use a maximum set of 130 financial statement predictors, so they cannot speak to the models' performance and computational costs in larger predictor sets. A third consideration is that combining technologies likely has advantages over using them in isolation,

[5] For example, studies find that tree-based methods can match or outperform earlier TFMs in other contexts, such as TabPFN applied within decision-tree partitions in bankruptcy prediction (Kostrzewa et al., 2025) and TabPFN 2.5 in clinical prediction (Shaktah et al., 2026). It is also plausible that tree-based models or Llama could outperform TabFM in our context if they were more extensively trained or fine-tuned.

for example, by using a large language model to construct features for a TFM or tree-based model.[6]

Our final caveat is that any pretrained model, including one pretrained on synthetic data such as TabFM, carries some potential for look-ahead bias (see footnote 2). A definitive test would evaluate forecasts of outcomes that were unknown when the model was publicly released, which is not yet possible for TabFM.[7] In a period of rapid AI development, however, a model is likely to be outdated by the time such a test can be run. Forecasters therefore face a trade-off between the risk of look-ahead bias and the cost of relying on older technology, and many may conclude that using the newer model is worth the risk. Given how Google describes TabFM's pretraining and the results of our memorization tests, it seems that look-ahead bias is unlikely to be a major concern for TabFM, but users should recognize that the risk is not zero.

Section 2 further explains TFMs and their connection to meta-learning. Section 3 presents our main comparisons of TabFM to an open-weight large language model and several tree-based models. Section 4 reports additional analyses of TabFM. Section 5 examines other TabFM settings and other TFMs, and Section 6 concludes.

## 2. Tabular Foundation Models

TFMs are pretrained models designed to transfer predictive capabilities across datasets organized into rows and columns. Models within this class differ in architecture, pretraining data, and adaptation procedures. We focus on implementations that use ICL, including TabPFN (Hollmann et al., 2023, 2025), TabICL (Qu et al., 2025), and TabFM (Kong & Das 2026; Kong et al., 2026). TabFM is our principal empirical implementation, and TabICLv2 (Qu et al., 2026) and TabPFN-3.5 (Jäger et al., 2026) provide comparisons in later tests.

---

[6] An extension of TabFM, TabFM-Auto, uses an LLM to construct additional predictors for use in TabFM (Fu et al., 2026). We do not evaluate TabFM-Auto due to concerns that the LLM introduces look-ahead bias by selecting predictors using future information it learned during pretraining.

[7] TabFM was released on June 30, 2026, after firms had announced their 2026 Q1 earnings. These earnings reveal part of the fiscal 2026 outcome, so a complete test would forecast fiscal 2027 earnings changes from fiscal 2026 financial statements. Firms report fiscal 2027 earnings in early 2028. By then, several newer generations of TFMs are likely to have been released. Footnote 10 discusses the same issue for Llama.

*2.1 How tabular foundation models learn and make predictions*

In a panel used in accounting and finance research, each row typically represents a firm-year, while columns contain financial statement items and other characteristics. For concreteness, we describe observations as firm-years, but the discussion applies equally to other units. A supervised prediction problem pairs these characteristics with an outcome that is known for the training sample but unknown for the test sample, which contains the observations requiring forecasts. For example, studies that contrast models often use current and lagged financial statement items to predict whether earnings increase in the following year.

Supervised learning estimates a forecasting relationship from labeled tabular data. Methods differ in how much of that relationship the researcher specifies. In a conventional statistical model, the researcher chooses the functional form, such as which predictors are used and in what form, and a fixed estimation rule converts predictors into predictions. For example, logistic regression uses maximum likelihood to compute the coefficients implied by the researcher's chosen specification. Tree-based machine learning methods instead learn the functional form from the data. A decision tree splits observations into groups using thresholds on the predictors, so it can capture nonlinear relationships and interactions that the researcher does not specify. Tree-based methods typically combine many trees, as in the random forests and gradient boosting models described in Section 3.2. The fitted relationship is then applied to new observations.

In tree-based methods, the researcher's choices therefore shift from the functional form, as in conventional statistical approaches, to settings that govern how the model learns. These settings, called hyperparameters, are chosen by the researcher rather than estimated from the data. They can include the number of trees, the depth of each tree, and the minimum number of observations in each final group. A configuration is one combination of hyperparameter values, and the grid is the set of candidate configurations. Researchers typically estimate the model under each configuration in the grid and select the one that performs best on a validation sample. We refer to this step, often called tuning, as the configuration search. Training consists of the configuration search and any re-estimation of the selected configuration. Predicting applies the trained model to the test sample.

Foundation models acquire reusable capabilities through broad initial training, called pretraining. Their parameters, which play a role similar to regression coefficients but are far more numerous, are estimated using a large collection of data. A language model typically learns to predict text from preceding text. The ICL-based TFMs discussed here learn by repeatedly predicting outcomes in synthetic tabular datasets. These datasets are generated from randomly drawn relationships between predictors and outcomes. For each generated dataset, the model receives some observations with their outcomes and predicts outcomes withheld for other observations. Prediction errors guide updates to the model's parameters, and repeating this exercise across many datasets teaches the model how to use examples from a new prediction problem. This process connects TFMs to meta-learning, which develops a reusable learning procedure through experience across tasks, often described as "learning to learn" (Hospedales et al., 2022). TFMs therefore go a step further than tree-based methods: the procedure for learning from the training sample is itself learned during pretraining, so their default settings are designed to be used without a configuration search.

After pretraining, researchers can adapt foundation models to a specific task through fine-tuning or in-context learning. Fine-tuning updates pretrained parameters using application-specific data. DF, for example, update a pretrained Llama model using financial statements paired with subsequent earnings-change outcomes.

Both LLMs and TFMs can use ICL, in which examples are supplied as model inputs, called the context, and the pretrained parameters remain fixed. Few-shot prompting in LLMs is a familiar application of ICL. In few-shot prompting, the prompt, or input supplied to the language model, contains a small number of worked examples followed by a new case (Brown et al., 2020). An LLM processes the examples as a sequence of tokens, which are the units used to represent text and numerical entries. For earnings prediction with a TFM, the researcher supplies financial statement predictors and known outcomes as the in-context training sample. The model processes these examples together with the predictors for the test-sample firm-years to produce a forecast. TFMs accept predictors and outcomes arranged in rows and columns.

### *2.2 Google Research's TabFM*[8]

Google Research's TabFM is a TFM that supports regression, which predicts numerical quantities, and classification, which predicts outcomes with up to ten categories. It combines large-scale synthetic pretraining with an architecture designed for tables.

Synthetic pretraining uses computer programs to generate datasets containing predictors and outcomes. Google reports that TabFM was pretrained on hundreds of millions of synthetic datasets generated using structural causal models incorporating varied random functions (Kong & Das 2026). A structural causal model generates variables through a system of directed relationships and random disturbances. For example, a generator could draw a first variable at random. It could then compute a second variable from the first using a randomly chosen function plus a random disturbance, and compute a third from the first two in the same way. Repeating this procedure with the same functions and new random draws produces the observations in a dataset. One of these variables is designated as the outcome, and the others serve as predictors. Varying the generating rules across datasets exposes the model to a range of relationships between predictors and outcomes, including nonlinear ones, and to different noise levels. The model learns to predict withheld outcomes from supplied examples, as described in Section 2.1. Because pretraining uses only generated data, actual firm outcomes cannot enter TabFM's parameters.

[8] Much of the technical discussion in this section applies to other TFMs.

TabFM's architecture processes a table through three operations. First, numerical entries are converted into vectors, which are lists of numbers representing information for the model. Attention, a mechanism that assigns data-dependent weights when combining information, is applied alternately across rows and columns. This allows the representation of a financial item to reflect both other items in the same observation and its pattern across observations. Second, the information for each row is compressed into a vector representing that observation. Working with these row vectors reduces the computational cost of the final step. Third, an in-context transformer, a model component built around attention operations, processes the compressed rows of the in-context training sample together with those of the test sample. It uses the training sample's known outcomes to produce the predictions. Because this step processes every training observation each time forecasts are produced, its running time and its use of GPU memory rise with the size of the in-context training sample.

These operations are relevant to financial statements because the information conveyed by one account (e.g., Net Income) can depend on its relationship with other accounts (e.g., Revenues and Expenses), and on how it compares with the same account at other firms.

Attention within a row relates items of the same firm-year, such that Revenues considered together with Expenses conveys different information from Revenues alone. Attention within a column relates each firm's value of an account to the values that other firms report. The in-context transformer then compares each test firm's overall profile with those of the in-context training firms.

By default, TabFM applies the model to 32 differently transformed copies of the data and averages the resulting predictions with equal weights. Each transformation is called an ensemble member. The members differ in how predictors are normalized, the order of the columns, and how the two outcome categories are coded. Every member therefore receives the same information, presented in a different way. Averaging across members makes the predictions less sensitive to these arbitrary choices. Each member requires a separate pass through the model, so running time rises with the number of members. Our main specification uses eight members, the default ensemble size of TabICLv2 and TabPFN-3.5. Google also describes an enhanced configuration, TabFM+, which keeps 32 members but adds constructed predictors (pairwise products and principal components of the original predictors) to half of them, weights the members using predictions for a validation portion of the training sample, and calibrates the resulting probabilities (Kong et al., 2026). The validation predictions require additional passes through the model, so TabFM+ takes longer than the default configuration. Section 5 examines different configurations of TabFM and TabFM+.

TabFM's pretraining used tables with at most 16,384 rows and 100 columns (Kong et al., 2026), but the current release accepts up to 500 predictors (Google Research 2026a). Our application, with about 56,000 in-context observations and 130 predictors, therefore relies on the model's ability to generalize to larger tables than those used in pretraining. We retain DF's 130 predictors for comparability and do not evaluate larger predictor sets. Although TabFM accepts more predictors than were used during pretraining, our results do not establish how its performance and computational requirements scale with further increases in the number of predictors.

*2.3 Applying tabular foundation models to accounting data*

Consider predicting, at the end of year $t$, whether a firm's earnings will increase in $t+1$. The researcher gives TabFM two samples of firm-years. The first is a training sample containing historical data through year $t$-1, in which each firm-year's financial statement items are paired with whether its earnings subsequently increased. To avoid look-ahead bias, all data in the training sample must already be publicly known by the forecast date. The second is the test sample containing firm-year data from year $t$, without future realizations in $t+1$. TabFM learns from the training sample and then produces a probability of an earnings increase for each observation in the test sample. Because all firms in our sample have December 31 fiscal year-ends, the outcomes of the year $t-1$ training observations (earnings changes in year $t$) are released in the same annual reports that supply the year $t$ test predictors.

In both samples, current and lagged financial statement items enter as separate columns, providing information about both the firm's recent financial history and relationships among its accounts. Because TabFM rescales each predictor internally, researchers can supply raw values without first standardizing them. Like tree-based models, TabFM ignores column names, so it does not know that a column contains net income or total assets. An LLM that reads financial statements as text, by contrast, sees the line-item labels.

The practical attraction of TabFM is that it natively handles tabular data and is pretrained to learn, so it can generate predictions without additional training. By contrast, tree-based models must be trained for each prediction task and, in settings such as earnings prediction, typically retrained each year. The computational costs of TabFM versus tree-based models therefore arise at different stages. A tree-based model incurs most of its cost in training, and applying the trained model to new firm-years takes seconds. TabFM has no training step, but it processes the in-context training sample each time it produces forecasts.

TabFM's pretraining raises two potential concerns about how it will perform in earnings prediction. First, each pretraining dataset consists of independent rows, whereas accounting panels contain repeated observations of the same firms. Second, each pretraining dataset is from a single generating process, whereas the relation between accounting values and future earnings changes over time. A related limitation is that TabFM, like tree-based models, ignores the order of rows in the training sample, so it cannot distinguish older observations from recent ones. Researchers can add time-related predictors as columns or restrict the training sample to recent years, although the first raises computational cost and the second shrinks the training sample. We do not explore these choices, although longer training windows improve performance in our setting despite possible shifts over time (Section 4.1).

## 3. Main Comparisons

### *3.1 Prediction task and sample*

Following prior research, our benchmark exercise predicts whether net income increases in the following fiscal year. An observation is classified as an increase when next-year net income exceeds current-year net income; all other observations are non-increases.

We use the dataset supplied by DF. It contains 62,796 firm-year observations, covering financial statement years 2005–2024 of 7,446 firms with December 31 fiscal year-end.[9] Each observation has 130 predictors, consisting of 26 income statement variables from the current year and prior two years, plus 26 balance sheet variables from the current and prior year. All items are unscaled amounts in millions of dollars, except basic and diluted earnings per share (six predictors), which are in dollars. See DF for further details of the sample construction. Table 1 reports the annual distribution of outcomes.

[9] Table 2 of DF's October 2026 draft lists 86,873 observations in Panel A, before limiting the sample to observations with necessary data. Panel B lists the 62,796 observations with available data, which is what we use.

DF avoid look-ahead bias from LLM pretraining by evaluating only earnings outcomes reported after Llama 3.2's release on September 25, 2024. Specifically, they predict earnings changes only from financial statements for 2023 and 2024, whose next-year earnings (i.e., years 2024 and 2025) were reported in 2025 and 2026.[10] We likewise limit our test years to 2023 and 2024 in the main comparisons that include DF's models, but later tests examine earlier years for models that cannot have memorized firms' outcomes in pretraining.

We evaluate models' performance based on AUC, which measures how effectively a model ranks observations with earnings increases above observations with non-increases. Equivalently, it is the probability that a randomly chosen firm-year with an earnings increase receives a higher predicted probability than a randomly chosen firm-year without one. An AUC of 50% represents uninformative ranking, while 100% represents perfect separation. AUCs are more informative than accuracy or F1 scores because they account for imbalances between the numbers of earnings increases and non-increases (Levy 2026).

### *3.2 Overview of comparison models*

We compare TabFM with the four models in DF: Llama, GBDT-DF, RF, and SGB. We use DF's predictions for SFT-Llama and GBDT-DF and re-estimate RF and SGB. We also add

[10] As of Llama 3.2's release date, firms' 2024 Q1 and Q2 results had been released, which could potentially contaminate prediction of fiscal 2024 earnings, the outcome for the 2023 test year. According to Meta, however, the knowledge cutoff date for the Llama 3.2 pretraining data is December 2023 (Meta 2024). DF also report that Llama outperforms the tree-based models in the 2024 test year alone.

GBDT-X, a faster re-estimation of GBDT-DF. Tree-based methods provide demanding benchmarks because they have performed well on tabular datasets relative to many neural-network alternatives (Grinsztajn et al., 2022). Below we briefly explain how each model is implemented, with further details in Appendix A:

- *TabFM.* Because TabFM does not require training, it needs no hold-out validation year. We therefore use all 18 years preceding each test year (2005–2022 for 2023 and 2006–2023 for 2024) as its in-context training sample.[11] Our main tests use TabFM 1.0.0 with eight ensemble members, and Section 5 examines other configurations.
- *Llama.* DF find that Meta's open-weight Llama-3.2-3B-Instruct can be fine-tuned for effective earnings prediction. They fine-tune Llama separately for each test year using a 17-year training window and a one-year validation window, from which they choose among eight data-augmentation configurations. DF run fine-tuning on a single NVIDIA H200 GPU. Given the cost of fine-tuning, we use the observation-level predicted probabilities from DF's "SFT-Llama" specification rather than fine-tuning the model ourselves.
- *GBDT-DF.* DF's first comparison model is a gradient-boosted decision tree (GBDT; Friedman 2001), which we label GBDT-DF. GBDT fits each new tree to the prediction errors that remain from the trees before it, meaning that trees must be fit one at a time. Every tree is fit to the full training sample. DF train GBDT-DF with the same training and validation windows as Llama. To give GBDT every opportunity to compete with Llama, DF search a broad grid of three hyperparameters (number of trees, learning rate, and maximum tree depth) with 1,760 combinations.[12] DF train each configuration separately on a single CPU core, running about 100 configurations in parallel on a server with two 64-core processors. The full search takes about two to three days per test year. Our analyses therefore use the observation-level predicted probabilities supplied by DF.[13]

[11] An untabulated robustness test shows that restricting TabFM to the same 17 estimation years used by the tree-based models reduces its AUC by 0.39 percentage points, and its AUC remains higher than those of all other models.
[12] See DF's Appendix D for details. In all tree-based models, we follow DF in not refitting the retained specification on the combined estimation and validation samples. Refitting would modestly increase training times.

- *RF.* A random forest (RF; Breiman 2001) averages independent trees grown on random subsamples of the data. The trees can be fit in parallel on an ordinary computer's CPUs. We implement RF in scikit-learn, using the same training and validation windows as our other tree-based models.
- *SGB.* Stochastic gradient boosting (SGB; Friedman 2002) follows the same sequential procedure as GBDT, but fits each tree to a random half of the training sample rather than the full sample. SGB's configuration search grows fewer and shallower trees (Appendix A.2), which makes it substantially faster to train. We implement SGB in XGBoost on a GPU.
- *GBDT-X.* GBDT-X re-estimates GBDT-DF using DF's 1,760-configuration grid and full-sample boosting, but with more efficient software and hardware. First, we use XGBoost on a GPU rather than scikit-learn on a CPU. Second, XGBoost groups each predictor into about 256 bins before searching for splits, whereas scikit-learn considers every value of each predictor. Binning is less computationally demanding, although possibly with an accuracy trade-off.

### *3.3 Main comparisons: performance*

Table 2A presents our main results for each model, pooling the results for 2023 and 2024 predictions. TabFM achieves an AUC of 69.53%, compared with 67.88% for Llama and 62.75% for GBDT-DF, and 63.51% to 63.78% for the other tree-based models. TabFM therefore ranks first, followed by Llama, while the tree-based models are close to one another.[14] Table 2A also reports accuracy relative to a 0.5 threshold for informational purposes and comparisons with DF.[15]

[13] DF use scikit-learn's GradientBoostingClassifier to train GBDT-DF over a 1,760-configuration grid. We re-estimated DF's selected configuration for each test year with their scikit-learn implementation, obtaining a pooled AUC of 62.51%, compared with their 62.75%.

Table 2B reports pairwise comparisons of AUCs between each of the models. All AUC comparisons use paired forecasts evaluated on the same observations, and we test AUC differences using a paired bootstrap with 10,000 replicates.[16] TabFM's performance advantages over all other models are highly statistically significant, as are Llama's advantages over the tree-based models, consistent with DF. Differences between tree-based models are generally statistically insignificant.

Figure 1A plots the ROC curves underlying these AUCs. Each curve shows the trade-off between correctly identifying earnings increases (true-positive rate) and misclassifying non-increases (false-positive rate) as the probability threshold varies. The AUC rankings in Table 2A therefore do not depend on the choice of threshold. Panel A shows that TabFM's curve lies above those of the other models across nearly the full range of false-positive rates, and Llama's lies above those of the tree-based models. The tree-based models' curves are nearly indistinguishable.

Table 2C and Figures 1B and 1C report AUCs separately for the two test years, and the results are similar. TabFM has the highest AUC in both years, followed by Llama, and then the tree-based models.[17]

[14] Our RF and SGB AUCs are within 0.04 percentage points of those reported by DF for the same models, indicating that our training and validation procedures closely resemble theirs.
[15] In our data, always predicting an earnings increase would be accurate 55.73% of the time.
[16] Each replicate draws firms with replacement and retains all test observations of each drawn firm, so the tests account for repeated observations from the same firm. Two-sided *p*-values equal twice the smaller of the bootstrap proportions of differences at or below zero and at or above zero, each computed with an add-one correction (Davison & Hinkley 1997). AUC differences are computed from unrounded values.
[17] When one year's model produces systematically higher probabilities, these cross-year comparisons can pull the pooled AUC below both annual AUCs.

### *3.4 Main comparisons: computing resources*

Table 2D reports the time and memory required for each model. All times are totals for both years. Because the models spend their time on different tasks, we separately report each model's time spent on training and predicting, and the sum of the two. Training time is incurred once for each training sample, whereas predicting time recurs each time forecasts are produced. The relative importance of the two therefore depends on how the model is used.

TabFM has no separate training step, so all of its time is in prediction. It takes 11.83 minutes, of which 2.33 minutes is consumed by loading the data and model, and preprocessing (untabulated). Our specification applies the model eight times per test year, one member at a time, which consumes the remaining 9.50 minutes.

Table 2D also reports the corresponding figures for Llama and GBDT-DF, provided by DF (Appendix A.3). Fine-tuning Llama under its eight candidate configurations takes about 108 hours on one NVIDIA H200 GPU, with peak GPU memory of up to 134 GiB. Its forecasts then take roughly 6 minutes. DF's GBDT-DF configuration search takes about two to three days per test year on a server with 128 CPU cores, running about 100 configurations in parallel. These figures are approximate and come from different hardware than ours, so they indicate orders of magnitude rather than precise comparisons.

Because these figures come from different hardware, we also compare the models' computing requirements in floating-point operations (FLOPs), the number of arithmetic operations a computation requires, which does not depend on the hardware used.[18] TabFM's forecasts for both test years require about $1.8 \times 10^{16}$ FLOPs. Fine-tuning Llama requires about

[18] We measure TabFM's FLOPs with PyTorch's FLOP counter. For Llama, we use the standard approximation of Kaplan et al. (2020), under which training requires about six FLOPs per model parameter for each token processed, and forecasting requires about two. Appendix A.3 provides details.

$1.2 \times 10^{20}$ FLOPs, roughly 6,700 times as many. Producing Llama's forecasts alone requires about $4 \times 10^{16}$ FLOPs, more than twice TabFM's total. There is no standard way to count FLOPs for tree-based models, so we report only their running times.

RF, SGB, and GBDT-X spend almost all of their time on training, whose configuration searches compare 64, 192, and 1,760 candidate configurations, respectively. GBDT-X takes the longest, at 140 minutes, because its search grows far more and deeper trees: 110 models of 8,000 trees each, with up to ten levels, compared with 192 models of 500 to 2,000 trees, with up to four levels, for SGB.[19] RF's long time despite its small grid reflects running on the CPU rather than the GPU and growing trees without a depth limit.

Table 2D also reports peak RAM and GPU memory usage. Peak RAM is modest for all models, at no more than about 14 GiB even with the full sample, which fits on many laptop computers. Peak GPU memory is minimal for SGB and GBDT-X, but TabFM's 24.3 GiB exceeds the capacity of most consumer GPUs and so typically requires a high-end consumer card, such as the one used here, although much less demanding than the 134 GiB from NVIDIA H200 used by DF to fine-tune Llama.

*3.5 Main comparisons: sensitivity to numerical perturbations*

As discussed, none of our models is pretrained on firms' outcomes. Still, for completeness, we apply the perturbation tests that prior studies use to detect memorization of firms' outcomes.

Levy (2026) uses least-significant-digit perturbations to investigate whether the apparent earnings-prediction ability of LLMs reflects look-ahead bias from memorization. Financial statement values can help an LLM recognize particular firm-years and recall subsequent earnings outcomes encountered during pretraining. Randomizing the last digit changes these identifying numerical patterns while preserving much of the financial information. If predictive performance depends on memorized associations, a one-digit perturbation should produce a substantial decline in AUC. For example, in Levy's tests, a one-digit perturbation reduces GPT-4's AUC from 57% to 48%, below the 50% of uninformative forecasts, while gradient-boosted trees decline from 65.2% to just 64.6%. DF apply perturbations to test whether their Llama model's performance reflects firm identification possibly learned during pretraining.

[19] As mentioned, we follow DF in examining 1,760 configurations, but researchers have discretion over the breadth of training searches. See Appendix A for further technical details of our implementations.

We test the sensitivity of our models to randomly replacing one, three, and five digits at the end of each test predictor's recorded value. Each condition uses ten random draws, and we report averages across draws. TabFM, RF, SGB, and GBDT-X receive identical perturbed test values, while all other model details remain unchanged. Our comparisons therefore evaluate how the four forecasting procedures respond to the same disturbance in the test predictors.

Figure 2 plots each model's AUC under each level of perturbation, together with the corresponding results from Levy and DF. Levy's sample differs from ours and DF use different perturbation draws, so their results provide only a qualitative benchmark. A one-digit perturbation removes all of GPT-4's skill in Levy's tests, but as expected, it only marginally affects Llama, TabFM, and the tree-based models. AUCs decline for all models as three and five digits are perturbed, which is expected as more information is removed. However, all models retain AUCs well above 50%. TabFM continues to have the highest AUC under each level of perturbation. Overall, the perturbation tests show no evidence of memorization in our models.

### *3.6 Main comparisons: discussion and conclusion*

Our analyses to this point indicate that TabFM compares favorably with Llama in both prediction accuracy and implementation cost in our setting. TabFM achieves an AUC of 69.53%, requires no training, and generates predictions in 11.83 minutes. Llama achieves an AUC of 67.88%, well above those of the tree-based models, but requires fine-tuning for about 108 hours on a data-center GPU.

TabFM also outperforms all the tree-based models in prediction accuracy, but the comparisons of implementation costs are more nuanced. TabFM requires no training, which can be a material advantage in cases where a user does not have the technical expertise to train tree-based models. However, TabFM's total time is longer than SGB's, and its predicting time is far longer than that of any tree-based model. In cases where a user will generate numerous predictions from the same model, tree-based models will be more efficient because they are trained just once and incremental predictions are nearly costless. Also, in cases where prediction time is essential, such as in event-based returns prediction and stock trading, tree-based models have a clear time advantage. Conversely, when the training sample is updated frequently, the tree-based models must repeat their configuration searches, whereas TabFM only needs the updated in-context training sample.

Among the tree-based models, GBDT-DF, RF, SGB, and GBDT-X perform similarly but have different computational requirements. RF might be the optimal choice for users who prefer to use a CPU, but SGB on a GPU is much faster. GBDT-X offers little advantage over the other tree-based models in our use-case. GBDT-DF has a slightly lower AUC than the other tree-based models and, in DF's scikit-learn implementation, takes substantially longer to train, indicating that its configuration procedures do not provide an advantage in our particular setting.

Section 4 examines other considerations that are relevant in practice.

## 4. Additional Analyses of TabFM

This section explores the performance of TabFM in various settings that might be relevant in research and practice. We contrast TabFM with RF, SGB, and GBDT-X. We do not include Llama and GBDT-DF because each analysis would require repeating costly fine-tuning or configuration search, and because Llama cannot be evaluated before 2023 without potential exposure to memorization-induced look-ahead bias. Unless otherwise noted, our analyses use the same variables, samples, and specifications as in Section 3.

*4.1 Performance by year*

We first evaluate variability in performance in earlier years. For each test year, RF, SGB, and GBDT-X use five training years followed by one validation year, repeating their configuration searches each year, while TabFM uses observations from the combined six-year period as its in-context training sample. Our data begin in 2005, so 2011 is our first test year. Each subsequent evaluation advances these periods by one year, for a total of 14 test years. The final two test years coincide with the main test sample but use shorter training windows.

The annual AUCs yield four observations. First, Figure 3 shows that TabFM outperforms the three tree-based models in every year except 2018, when TabFM's AUC is slightly lower than SGB's. Across all years, Table 3A reports that TabFM has a pooled AUC of 69.52%, compared with 64.82% for SGB, the next highest, and Table 3B reports that the paired differences in AUCs are highly statistically significant.

Second, TabFM's higher AUC does not come with greater annual variation. Table 3A reports that the standard deviation of TabFM's annual AUCs is 4.39 percentage points, the same as GBDT-X's and lower than those of RF and SGB. TabFM's advantage is also stable. The standard deviations of its annual advantages over the tree-based models, 1.92 to 2.32 points, are less than half its average annual advantages of 4.88 to 5.33 points.

Third, shorter training windows reduce the performance of all models. With the 18-year windows in our main tests, the 2023 AUCs are 70.18% for TabFM and 62.72% to 63.63% for the tree-based models (Table 2C). With six-year windows, they are 67.76% for TabFM and 59.50% to 60.80% for the tree-based models (Figure 3).

Fourth, all models' annual AUCs vary substantially across years. TabFM's AUC ranges from 62.82% in 2016 to 78.86% in 2017, a swing of 16 percentage points between consecutive years, and the tree-based models vary similarly. The four models' annual AUCs have pairwise correlations of 0.89 to 0.99 (untabulated), indicating that annual variability reflects how predictable earnings changes are in a given year rather than model-specific performance. An evaluation based on one or two test years therefore provides an incomplete picture of a model's predictive ability. This caveat applies to any evaluation limited to a few test years, including our main comparisons, which are constrained by Llama's release date.

*4.2 Sensitivity to omitting features*

Financial statements contain overlapping, correlated variables, and many financial statement items are linked by accounting identities through which components sum to subtotals. As a result, the information in any single item is often available elsewhere in the statements, and its measured importance as a predictor depends on how well a model can recover that information from the remaining items.

We evaluate how performance changes when predictors are removed, rerunning each model's training and prediction procedure on the reduced set. Table 4 reports five omission conditions:

- The first condition removes the five predictors that DF identify as most important for Llama, all of which are current-year income measures: 1) net income; 2) income before extraordinary items and noncontrolling interest; 3) income before extraordinary items; 4) income before extraordinary items for common stocks; and 5) income before extraordinary items adjusted for common stock equivalents. Table 4 shows dropping these measures has an immaterial effect on all four models' AUCs. TabFM, for example, drops from an AUC of 69.53% to 69.48%, and the difference is statistically insignificant.[20]

- The second condition also removes the values of the same measures from the two preceding years, leaving 115 predictors remaining. TabFM's AUC declines by 1.09 points, and those of RF and SGB by 0.91 and 0.71 points, all statistically significant.
- The third condition removes the 20 most important predictors from DF, which are a mix of current and prior-year measures. TabFM's AUC drops by 1.31 points, compared with 1.95 to 2.50 points for the tree-based models.
- The fourth condition removes all predictors from the $t-2$ income statement and $t-1$ balance sheet, such that just one year's balance sheet and two years' income statements remain. The AUCs of TabFM, SGB, and GBDT-X are virtually unchanged, while RF's rises by 0.95 points.
- The fifth condition removes all predictors from the $t-2$ income statement and both years' balance sheets, such that just two income statements remain (52 predictors total). The AUCs for TabFM and RF are unchanged, while those of SGB and GBDT-X drop slightly.

Overall, Table 4 shows that all four models are remarkably resilient to removing potentially important variables, and even to removing the balance sheet altogether. Removing variables also has little effect on the relative performance across models; in all cases, TabFM

[20] Our inference differs from DF's permutation-importance analysis, in which shuffling current-year net income across firms within the same year reduces Llama's AUC by 9.67 percentage points (DF, Table 5). The two approaches answer different questions: permutation importance measures how much a fitted model relies on a predictor, whereas removing a predictor and refitting the model measures how much information the predictor adds beyond the others (Fisher et al., 2019; Hooker et al., 2021). Because net income is closely related to several other income measures, a model can rely heavily on it even though little of its information is unique.

outperforms the tree-based models.

*4.3 Performance with smaller training samples*

A practical question is how forecasting performance changes when fewer training observations are available. A researcher studying a single industry, for example, may have only a few thousand firm-years, and a firm predicting accounts receivable write-offs may have data on only a few hundred customers. To examine this question, we conduct two sets of analyses.

First, we randomly select 80%, 60%, 40%, 20%, 10%, or 5% of the training observations within each year, without replacement. The samples are nested, so within each draw every smaller sample is a subset of the larger ones. We then repeat the training and test procedures from our main analysis, using each reduced sample. The tree-based models' validation year is not reduced, whereas TabFM's in-context training sample includes that year at the reduced fraction. This design favors the tree-based models. Each model then forecasts the same 6,569 test observations from 2023 and 2024. We repeat this procedure with ten random seeds and report the average of each performance measure across the ten draws.

Table 5A reports averages across the ten draws. TabFM's in-context training sample falls from about 56,200 firm-years per test year in the full sample to about 2,800 in the 5% sample. TabFM has the highest average AUC at every fraction. Its AUC declines from 69.53% to 64.47% at 5%, which still exceeds the full-sample AUCs of the tree-based models. This 5.06-point decline is larger than those of the tree-based models, which range from 3.54 to 4.01 points, likely because TabFM starts from a higher baseline. Measured as skill lost, the share of above-chance AUC that is lost, TabFM's decline of 25.90% is similar to SGB's 25.70% and smaller than those of RF (29.68%) and GBDT-X (29.09%).

Table 5B reports each model's running time and peak memory use alongside its AUC.

TabFM's time falls by 89.1% between the full and 5% samples, and RF's falls by 94.4%. The times of SGB and GBDT-X fall by about half, likely because much of their cost does not depend on the number of observations.[21] TabFM is the fastest model at 60% of the sample and below. At 5%, it takes 1.28 minutes, compared with 3.91 for SGB. TabFM's peak GPU memory falls by 68%, although at 7.85 GiB it remains well above theirs.

Second, Tables 5C and 5D report a test that randomly retains subsets of firms instead of observations for any firm. For each test year, we randomly draw 500 firms and use their observations over the preceding six years, about 2,200 to 2,300 firm-years, as the training sample. The tree-based models use the first five years for estimation and the sixth for validation, so their validation samples also shrink. We then retain nested subsets of 400, 300, 200, and 100 of these firms. Every condition is evaluated on the same 1,000 test-year observations, and we average results across ten draws. With 500 firms, TabFM's AUC of 62.92% exceeds the tree-based models' AUCs by 4.9 to 6.2 points. The results are similar with 100 firms. Between 500 and 100 firms, TabFM loses 26.39% of its skill, compared with 18.97% to 29.83% for the tree-based models. TabFM is also the fastest model at every firm count, taking 0.65 to 0.95 minutes, compared with at least 2.18 minutes for the tree-based models.

Overall, all models retain meaningful predictive ability as the training sample shrinks, and TabFM retains the highest AUC throughout. Between the full and 5% samples, TabFM's running time falls by 89% and its peak GPU memory by 68%, whereas SGB's and GBDT-X's running times fall by only about half.

## 5. Analysis of Other TabFM Settings and Other Tabular Foundation Models

[21] RF grows fully developed trees on the CPU, so the cost of each tree rises with the number of observations. SGB and GBDT-X grow depth-limited trees on the GPU, where much of each tree's cost is likely fixed, and their configuration searches grow the same number of trees at every fraction.

This section compares our main TabFM results with those of two other TFMs and with alternative configurations of TabFM. The two other TFMs, TabICLv2 and TabPFN-3.5, are pretrained and applied in much the same way as TabFM (Qu et al., 2026; Jäger et al., 2026). They differ from TabFM mainly in model sizes and architecture, which affect running time and memory use, and in their synthetic pretraining data. Within TabFM, we evaluate the 32-member ensemble and TabFM+, as described in Section 2.2. Finally, we compare TabFM version 1.0.0, used to this point, with the recently released version 1.1.0, which has about one-quarter as many parameters. All specifications receive the same predictors, training samples, and test observations as in Section 3. Appendix A.1 describes their implementation settings.

Table 6A shows that the AUCs of all the TFM specifications are within two percentage points of one another. Our main TabFM specification has the highest AUC, at 69.53%, and the other specifications range from 67.57% to 69.48%. Table 6B shows that the main specification's lead is statistically significant over every other specification except TabFM with 32 members ($p = 0.44$). All three TabFM 1.0.0 specifications significantly outperform Llama, whereas TabICLv2, TabPFN-3.5, and the three TabFM 1.1.0 specifications do not differ significantly from it. The main TabFM specification also has the highest AUC of the three TFMs in each test year: 70.18% in 2023 and 68.69% in 2024, compared with 67.71% and 67.39% for TabICLv2 and 69.85% and 67.05% for TabPFN-3.5 (untabulated). Finally, every TFM specification significantly outperforms each tree-based model.

Computational requirements differ substantially across the TFMs. TabICLv2 and TabPFN-3.5 produce forecasts in about 1 and 3 minutes, respectively, compared with about 12 minutes for our main TabFM specification. Within TabFM, running time rises with the number of ensemble members. The 32-member specification takes about 41 minutes, and TabFM+ takes about 83 minutes. Each TabFM 1.1.0 specification takes about half the time of its 1.0.0 counterpart, consistent with 1.1.0 using a smaller underlying model. The ranking differs for GPU memory. TabICLv2, the fastest model, has the highest peak GPU memory, at 30.5 GiB, compared with 24.3 GiB for TabFM and 3.8 GiB for TabPFN-3.5. These differences partly reflect how each package schedules its computations.[22]

We hesitate to draw strong inferences from Table 6, because current TFMs are early versions of what will likely emerge over the next few years. Still, faster prediction tends to come with lower AUC. TabICLv2 and TabPFN-3.5 are faster than TabFM 1.0.0 but less accurate, and TabFM 1.1.0 roughly halves 1.0.0's running time and GPU memory but lowers AUC. This trade-off matters for the comparison with Llama: TabFM 1.0.0 significantly outperforms Llama, whereas TabFM 1.1.0 does not. The pattern does not always hold, however. Moving from eight to 32 members, or to TabFM+, multiplies running time without improving AUC. Time is also not the only cost. TabICLv2 is the fastest model but requires the most GPU memory, while TabPFN-3.5 comes within one percentage point of TabFM's AUC at a fraction of its GPU memory.

## 6. Conclusion

This technical note explains TFMs and their application to accounting panel data. These models combine native handling of tabular data with predictive capabilities developed through pretraining across synthetic datasets. In-context learning applies those capabilities using labeled training observations supplied for the researcher's forecasting problem. TFMs pretrained only on synthetic data, such as TabFM, have no inherent accounting knowledge and cannot have memorized firms' outcomes.

Our central finding is that TabFM, the best-performing TFM we test, predicts earnings changes more accurately than both a fine-tuned large language model and several tree-based models. TabFM's advantage over the tree-based models holds across years, subsets of predictors, and training-sample sizes. It also requires no configuration search. TabFM compares favorably with the large language model on cost, whereas its comparison with tree-based models involves a trade-off that depends on how often forecasts are produced and how often the training sample changes.

[22] TabICLv2 processes its ensemble members together, which is fast but memory-intensive, whereas TabFM processes its members one at a time.

Our inferences have two important caveats. First, our results come from one benchmark task, one sample, and one set of predictors, and no single forecasting technology is likely to outperform another universally. Second, the assumption that TabFM has not memorized firms' outcomes relies on the developers' reports of how the model was pretrained, although our tests show no sign of memorization. Synthetic pretraining also does not rule out broader forms of look-ahead bias, such as design choices informed by real data, which apply to any forecasting method.

Several directions seem promising. First, researchers can evaluate TFMs using other measures of performance and cost, and in other settings of interest to accountants, such as fraud detection and credit risk assessment. Second, translating differences in AUC into economic value would show what a more accurate model is worth to its users, and whether that value exceeds the cost of producing its forecasts. Finally, much of the earnings-forecasting literature aims not only to forecast earnings but also to understand the role that accounting numbers play in the forecasting process (Monahan 2018). Our analysis, by contrast, treats TabFM as a black box. Because TFMs learn how financial statements relate to future earnings from the data alone, without accounting knowledge or a researcher-specified model, examining what they learn could improve our understanding of the underlying properties of earnings processes.

## References

Breiman, L. 2001. Random forests. *Machine Learning* 45 (1): 5–32.

Brown, T. B., B. Mann, N. Ryder, M. Subbiah, J. Kaplan, P. Dhariwal, A. Neelakantan, P. Shyam, G. Sastry, A. Askell, S. Agarwal, A. Herbert-Voss, G. Krueger, T. Henighan, R. Child, A. Ramesh, D. M. Ziegler, J. Wu, C. Winter, C. Hesse, M. Chen, E. Sigler, M. Litwin, S. Gray, B. Chess, J. Clark, C. Berner, S. McCandlish, A. Radford, I. Sutskever, and D. Amodei. 2020. Language models are few-shot learners. *Advances in Neural Information Processing Systems* 33: 1877–1901.

Chen, X., Y. H. Cho, Y. Dou, and B. Lev. 2022. Predicting future earnings changes using machine learning and detailed financial data. *Journal of Accounting Research* 60 (2): 467–515.

Davison, A. C., and D. V. Hinkley. 1997. *Bootstrap Methods and Their Application*. Cambridge: Cambridge University Press.

Dou, Y., and Q. Fu. 2026. Supervised fine-tuning of large language models on financial statements to predict future earnings changes. Working paper, New York University. SSRN working paper 6484841.

Fisher, A., C. Rudin, and F. Dominici. 2019. All models are wrong, but many are useful: Learning a variable's importance by studying an entire class of prediction models simultaneously. *Journal of Machine Learning Research* 20 (177): 1–81.

Freeman, R. N., J. A. Ohlson, and S. H. Penman. 1982. Book rate-of-return and prediction of earnings changes: An empirical investigation. *Journal of Accounting Research* 20 (2): 639–653.

Friedman, J. H. 2001. Greedy function approximation: A gradient boosting machine. *The Annals of Statistics* 29 (5): 1189–1232.

Friedman, J. H. 2002. Stochastic gradient boosting. *Computational Statistics & Data Analysis* 38 (4): 367–378.

Fu, D., H. Su, R. Sen, T. Narayan, S. Sanghavi, A. Das, and W. Kong. 2026. TabFM-Auto: Self-evolving pipelines for tabular foundation models. Working paper, arXiv:2609.37989.

Google Research. 2026a. TabFM: Tabular foundation models. Software, implementation documentation, and TabFM 1.0.0 model card. GitHub repository: google-research/tabfm; Hugging Face model repository: google/tabfm-1.0.0-pytorch. Accessed September 30, 2026.

Google Research. 2026b. TabFM 1.1.0 model card. Hugging Face model repository: google/tabfm-1.1.0-pytorch. Accessed September 30, 2026.

Grattafiori, A., A. Dubey, A. Jauhri, A. Pandey, A. Kadian, et al. 2024. The Llama 3 herd of models. Working paper, arXiv:2407.21783.

Grinsztajn, L., E. Oyallon, and G. Varoquaux. 2022. Why do tree-based models still outperform deep learning on typical tabular data? *Advances in Neural Information Processing Systems* 35: 507–520.

Hollmann, N., S. Müller, K. Eggensperger, and F. Hutter. 2023. TabPFN: A transformer that solves small tabular classification problems in a second. *International Conference on Learning Representations*.

Hollmann, N., S. Müller, L. Purucker, A. Krishnakumar, M. Körfer, S. B. Hoo, R. T. Schirrmeister, and F. Hutter. 2025. Accurate predictions on small data with a tabular foundation model. *Nature* 637: 319–326.

Hooker, G., L. Mentch, and S. Zhou. 2021. Unrestricted permutation forces extrapolation: Variable importance requires at least one more model, or there is no free variable importance. *Statistics and Computing* 31: 82.

Hospedales, T., A. Antoniou, P. Micaelli, and A. Storkey. 2022. Meta-learning in neural networks: A survey. *IEEE Transactions on Pattern Analysis and Machine Intelligence* 44 (9): 5149–5169.

Jäger, B., N. Erickson, L. Grinsztajn, et al. 2026. TabPFN-3.5: Technical report. arXiv:2609.17895.

Kaplan, J., S. McCandlish, T. Henighan, T. B. Brown, B. Chess, R. Child, S. Gray, A. Radford, J. Wu, and D. Amodei. 2020. Scaling laws for neural language models. Working paper, arXiv:2001.08361.

Kong, W., and A. Das. 2026. Introducing TabFM: A zero-shot foundation model for tabular data. *Google Research Blog*, June 30.

Kong, W., E. Louidor Ilan, S. Nie, T. Narayan, R. Sen, Y. Zhou, D. Fu, S. Oymak, and A. Das. 2026.

TabFM: A zero-shot foundation model for tabular data. Working paper, arXiv:2609.37959.
Kostrzewa, M., O. Furman, R. Furman, S. Tomczak, and M. Zięba. 2025. Are foundation models useful for bankruptcy prediction? Working paper. NeurIPS 2025 Workshop on Generative AI in Finance. arXiv:2511.16375.
Levy, B. 2026. Caution ahead: Numerical reasoning and look-ahead bias in AI models. *Journal of Accounting Research* 64 (3): 1139–1188.
Meta. 2024. Llama 3.2 model card. GitHub repository: https://github.com/meta-llama/llama-models/blob/main/models/llama3_2/MODEL_CARD.md.
Monahan, S. J. 2018. Financial statement analysis and earnings forecasting. *Foundations and Trends in Accounting* 12 (2): 105–215.
Müller, S., N. Hollmann, S. Pineda Arango, J. Grabocka, and F. Hutter. 2022. Transformers can do Bayesian inference. *International Conference on Learning Representations*.
Ou, J. A., and S. H. Penman. 1989. Financial statement analysis and the prediction of stock returns. *Journal of Accounting and Economics* 11 (4): 295–329.
Qu, J., D. Holzmüller, G. Varoquaux, and M. Le Morvan. 2025. TabICL: A tabular foundation model for in-context learning on large data. *Proceedings of the 42nd International Conference on Machine Learning*, Proceedings of Machine Learning Research 267: 50817–50847.
Qu, J., D. Holzmüller, G. Varoquaux, and M. Le Morvan. 2026. TabICLv2: A better, faster, scalable, and open tabular foundation model. *Proceedings of the 43rd International Conference on Machine Learning*, Proceedings of Machine Learning Research 306: 102477–102555.
Shaktah, L. A., M. Gustav, T. Lenz, J. Liang, L. Hilgers, Z. I. Carrero, and J. N. Kather. 2026. Established machine learning matches tabular foundation models in clinical predictions. *BMC Medical Informatics and Decision Making* 26: 364.

## Appendix A. Forecasting implementation

### *A.1 Tabular foundation model implementations*

#### *A.1.1 TabFM*

We use Google's TabFM 1.0.0 PyTorch classification checkpoint, obtained from the Hugging Face model repository google/tabfm-1.0.0-pytorch at revision 77cb9cc1b4fd3a9c77fbb9552c218200bb4dab83, with software version tabfm 1.0.1. The computational backend is PyTorch 2.8.0 with CUDA 12.8, using a single CUDA-enabled GPU. The same pretrained checkpoint is used throughout the evaluation, except for the TabFM 1.1.0 results in Table 6. These use the 1.1.0 classification checkpoint from the Hugging Face repository google/tabfm-1.1.0-pytorch at revision d7cfd3b8d184d7c8fed4210aeba9e1c967d45f2a, with software version tabfm 1.0.1, under the eight-member, 32-member, and TabFM+ configurations.

The released model card describes three column-attention blocks, three row-attention blocks, and a 24-block in-context transformer, with an embedding dimension of 256. A block is a repeated group of network operations, while the embedding dimension describes the number of numerical components used in the relevant internal representation (Google Research 2026a, 2026b).

TabFM's default ensemble combines prediction scores from 32 applications of the same pretrained model to differently transformed versions of the data. The members differ in predictor normalization, column order, and encoding of the two outcome categories. Predictors are standardized, with half of the members also applying a Yeo–Johnson power transformation. The default outlier treatment uses a threshold of four standard deviations. The raw prediction scores, or logits, are averaged with equal weights and converted into probabilities using a softmax temperature of 0.9. No probability calibration is applied. Kong et al. (2026) also describe an enhanced configuration, TabFM+. Like the default, it uses 32 members, but it adds predictors constructed from pairwise products and principal components to half of them. It also weights the members using predictions for a validation portion of the training sample, with the estimated weights shrunk 25% toward equal weights. It averages probabilities rather than logits and calibrates them using Platt scaling. TabFM's default setting passes the ensemble members through the model one at a time, so running time increases with the number of members. Our main specification uses eight ensemble members, the default ensemble size of TabICLv2 and TabPFN-3.5, holding all other settings fixed. As robustness checks, we also apply TabFM with its default 32 members, and TabFM+ using the software's ensemble preset (TabFMClassifier.ensemble), to the principal test sample. Section 5 and Table 6 report the results. All other reported TabFM results use eight members.

We retain the default settings of max_num_rows=None and max_num_features=500, so the software uses all supplied training observations and all 130 predictors rather than a subsample. Each ensemble member uses all training observations selected for the relevant forecast year and experimental condition. This configuration is held fixed across the principal, smaller-sample, omission, perturbation, firm-count, and annual analyses.

Ensemble generation and the Python, NumPy, and PyTorch random-number generators in the application scripts use fixed seeds. Test predictions are processed in batches of up to 9,999 observations, so all test observations for a year are processed in a single batch. Each forecast year therefore requires one pass per ensemble member over the in-context training sample and the test observations. Because each test observation attends only to the in-context training sample, the batch size affects running time but not the predictions (Kong et al., 2026).

We do not tune TabFM's configuration, and the pretrained model parameters remain fixed during application. Preprocessing parameters are estimated from the designated in-context training sample. TabFM uses only the numerical values of the predictors and ignores column names. The reported probability corresponds to the earnings-increase category, with the model's class ordering checked against the outcome coding.

### *A.1.2 TabICLv2 and TabPFN-3.5 robustness implementations*

The robustness analyses use TabICLv2 (Qu et al., 2026) and TabPFN-3.5 (Jäger et al., 2026). For TabICLv2, the checkpoint is tabicl-classifier-v2-20260212.ckpt, and the software version is tabicl 2.2.0. For TabPFN-3.5, the checkpoint is tabpfn-v3.5-20260909.safetensors, and the software version is tabpfn 9.0.0.

Both models run on a single CUDA-enabled GPU with a fixed random seed. We retain the default prediction configurations of their respective software packages. This configuration applies TabICLv2's default numerical preprocessing with an eight-member ensemble, and TabPFN-3.5's default preprocessing pipeline with an eight-member ensemble. The application scripts pass each full year-specific training window to the models without subsampling rows, and neither package subsamples rows internally at our sample sizes. Test predictions are processed in batches of up to 9,999 observations, matching TabFM, so all test observations for a year are processed together. Each model's checkpoint and application rules are held fixed across the two principal test years, subject to the training observations available in each year.

### *A.2 RF, SGB, and GBDT-X implementations*

The RF implementation uses scikit-learn 1.9.1, and the SGB and GBDT-X implementations use XGBoost 3.4.1. Configuration selection is based on validation AUC. The candidate configurations are specified before evaluating the test sample.

RF is implemented as a bagged ensemble of decision trees using scikit-learn's BaggingClassifier. Each tree is fitted to a random half of the training observations, drawn without replacement, and considers the square root of the number of predictors at each split. Trees use the Gini criterion and no maximum depth. The grid varies the number of trees from 500 to 2,000 in steps of 100 and the minimum terminal-node size from one to four, yielding 64 candidate configurations. The selected configurations use 1,000 trees for the 2023 test year and 600 trees for 2024, both with a minimum terminal-node size of one.

SGB uses XGBoost's tree booster with histogram-based split finding on the GPU. Each boosting stage, which adds one tree, uses a random half of the training observations. We set XGBoost's minimum child weight (min_child_weight) to 2.5, set the L2 penalty on leaf weights (reg_lambda) to zero, and minimize log loss. The grid varies the number of boosting stages from 500 to 2,000 in steps of 100, the learning rate over 0.005, 0.01, and 0.05, and the maximum tree depth from one to four, yielding 192 candidate configurations. The selected configurations use 1,700 stages, a learning rate of 0.05, and a maximum depth of four for 2023, and 2,000 stages, a learning rate of 0.05, and a maximum depth of three for 2024. The grid follows Chen et al. (2022), who implement SGB in R's gbm package.

GBDT-X re-estimates GBDT-DF in XGBoost. GBDT-DF uses scikit-learn's GradientBoostingClassifier and, following Levy (2026), fits each boosting stage to the full training sample (DF, Appendix D). Its grid varies the number of boosting stages from 500 to 8,000 in steps of 500, the learning rate over 0.001, 0.005, 0.01, 0.05, 0.1, 0.15, 0.2, 0.25, 0.3, 0.35, and 0.4, and the maximum tree depth from one to ten, yielding 1,760 candidate configurations. DF select 1,500 stages, a learning rate of 0.25, and a maximum depth of nine for 2023, and 4,500 stages, a learning rate of 0.05, and a maximum depth of five for 2024. GBDT-X searches the same grid and, unlike SGB, fits each boosting stage to the full training sample. To align XGBoost with scikit-learn's gradient boosting, we set the L2 penalty on leaf weights (reg_lambda), the L1 penalty (reg_alpha), the minimum child weight (min_child_weight), and the minimum loss reduction for a split (gamma) to zero. Trees grow depth-wise with no limit on the number of leaves, and every tree uses all observations and all predictors.

Unlike scikit-learn, XGBoost's histogram method groups each predictor into about 256 bins and considers splits only at bin boundaries, and it weights observations differently when choosing splits, so its results do not match DF's exactly. For each pair of learning rate and depth, we fit one model with 8,000 stages and evaluate the 16 numbers of stages from that fit. Because boosting adds trees without changing earlier ones, the first *n* stages of that model are identical to a model fitted with *n* stages, so this gives the same results as fitting each configuration separately. The selected configurations are 1,000 stages, a learning rate of 0.25, and a maximum depth of five for 2023, and 2,000 stages, a learning rate of 0.05, and a maximum depth of seven for 2024.

RF, SGB, and GBDT-X use a fixed random seed. The selected tree-based models are applied to the test observations without refitting on the combined training and validation samples.

*A.3 Computational measurement*

Computational requirements for TabFM, TabICLv2, TabPFN-3.5, RF, SGB, and GBDT-X are measured on a Linux server (kernel 5.15, x86-64) with an Intel Xeon Platinum 8352V CPU (2.10 GHz) and 128 logical processors, one NVIDIA GeForce RTX 4080 SUPER GPU provided through a cloud computing platform with its memory expanded to approximately 32 GiB (retail versions have 16 GiB), and 503.5 GiB of system memory, using Python 3.12.3, NumPy 2.3.2, pandas 3.0.6, and PyTorch 2.8.0. TabICLv2's peak GPU memory of 30.5 GiB would therefore exceed the capacity of a retail card. Our runs are allocated 16 of the 128 logical processors, each of which executes one thread, or stream of computation, at a time. RF runs on the CPU only, fitting up to eight trees in parallel. SGB and GBDT-X use four CPU threads together with the GPU. TabFM, TabICLv2, and TabPFN-3.5 perform their model computations on the GPU, using their packages' default CPU threading.

Each reported time and memory measurement comes from a single execution of the relevant model and experimental condition, except in the reduced-sample conditions of the smaller-sample analysis and the firm-count analysis, where measurements are averaged across the ten draws. We divide each run's elapsed time into training and predicting. Training is the time RF, SGB, and GBDT-X spend estimating each candidate configuration on the estimation years and computing its AUC on the validation year. Because the selected configuration is not re-estimated, this is their full training time. The TFMs have no training step. Predicting is the remaining time. It covers starting the process, reading the data, and producing probabilities for the test observations. For the TFMs, it also includes loading the pretrained model, fitting the preprocessing steps to the in-context training sample, and processing that sample. GBDT-X processes the two test years concurrently on the same GPU. Its reported time is the sum of the elapsed times for the two years, which may slightly overstate the time required to run them in sequence because the concurrent runs share the GPU and processors. Its peak RAM is the larger of the two years' peaks, each measured for that year's process alone. Its peak GPU memory is measured device-wide while both years run, so it includes both processes and is higher than the memory needed to run one year at a time.

System memory (RAM) is sampled at 0.05-second intervals using psutil 7.0.0, summing memory usage across the Python process and its child processes. GPU memory is sampled at one-second intervals using nvidia-smi. The tables report peak RAM use and peak GPU memory. Peak GPU memory is the device-wide maximum during the run, including the CUDA context, less the memory already in use on the device when the run starts. This measure is comparable across the TFMs and SGB. RF does not use the GPU. For GBDT-X, RAM is measured for each year's process and GPU memory for both concurrent processes together, as described above, so its memory figures are not directly comparable with those of the other models.

The computational figures for Llama and GBDT-DF in Table 2D were provided by DF (Y. Dou, personal communication, October 2026). DF fine-tune Llama-3.2-3B-Instruct on a single NVIDIA H200 GPU with about 140 GiB of memory. For each test year, fine-tuning the eight data-augmentation configurations takes about 54 GPU-hours. Peak GPU memory during fine-tuning is 134 GiB for the 2023 test year and 124 GiB for the 2024 test year. DF produce Llama's forecasts on the same GPU with a batch size of one, which takes about two to four minutes per test year. DF run the GBDT-DF configuration search on a server with two 64-core AMD EPYC 9534 processors and up to 1.2 TB of RAM. They train each of the 1,760 configurations independently on a single core, with about 100 configurations running in parallel. DF estimate that the search requires about 34 to 44 hours per test year, but the actual elapsed time is about two to three days per test year because the search has to wait for its slowest configurations to finish. Peak RAM for either model and the time required to produce GBDT-DF's forecasts are not available.

We measure TabFM's FLOPs with PyTorch's FLOP counter in a separate run of the main specification, which gives $1.82\times10^{16}$ FLOPs for both test years. Llama-3.2-3B has about 2.8 billion parameters outside its embedding layer. DF's example prompt (DF, Appendix B) contains about 1,100 tokens under Llama's tokenizer, and we treat this as a typical firm-year. DF's random removal of 20% or 40% of line items reduces this to about 860 tokens in training. With about 53,000 training firm-years per test year, 10 epochs, and eight data-augmentation configurations, fine-tuning processes about 7.3 billion tokens over the two test years, or about $1.2\times10^{20}$ FLOPs. Evaluating the eight configurations on the two validation years (about 6,500 firm-years) processes about 58 million tokens, or about $3\times10^{17}$ FLOPs. Producing forecasts for the 6,569 test firm-years processes about 7 million tokens, or about $4\times10^{16}$ FLOPs, the figure reported in Section 3.4. As a check on our approximation, DF's 54 H200 GPU-hours per test year imply that the GPU sustained about one-third of its peak speed, a plausible rate, somewhat below the 38–43% that Meta reports for pretraining its largest Llama 3 model (Grattafiori et al. 2024).

## Figure 1. Receiver Operating Characteristic (ROC) Curves

This figure compares ROC curves for TabFM, Llama, GBDT-DF, RF, SGB, and GBDT-X. Panel A shows the pooled sample based on 2023 and 2024 financial statements, and Panels B and C show the 2023 and 2024 financial statement years separately. The vertical axis shows the true-positive rate, the proportion of actual earnings increases correctly identified. The horizontal axis shows the false-positive rate, the proportion of non-increases incorrectly classified as increases. Each curve traces these two rates as the probability threshold for classifying an observation as an increase varies. The dashed diagonal represents uninformative forecasts, and the legend reports each model's AUC. See Section 3.3 for discussion.

*Panel A: Pooled, 2023–2024*

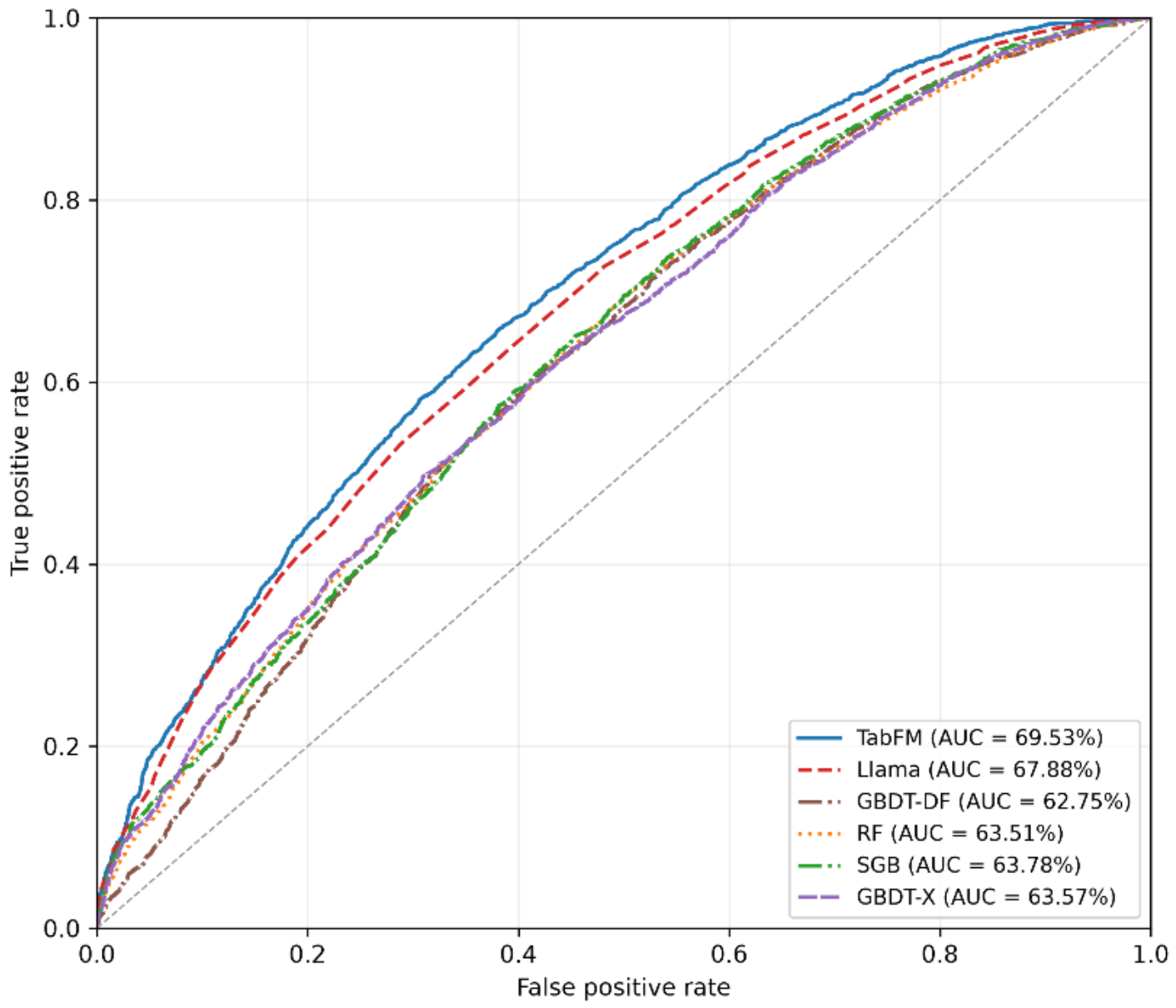

*Panel B: 2023*

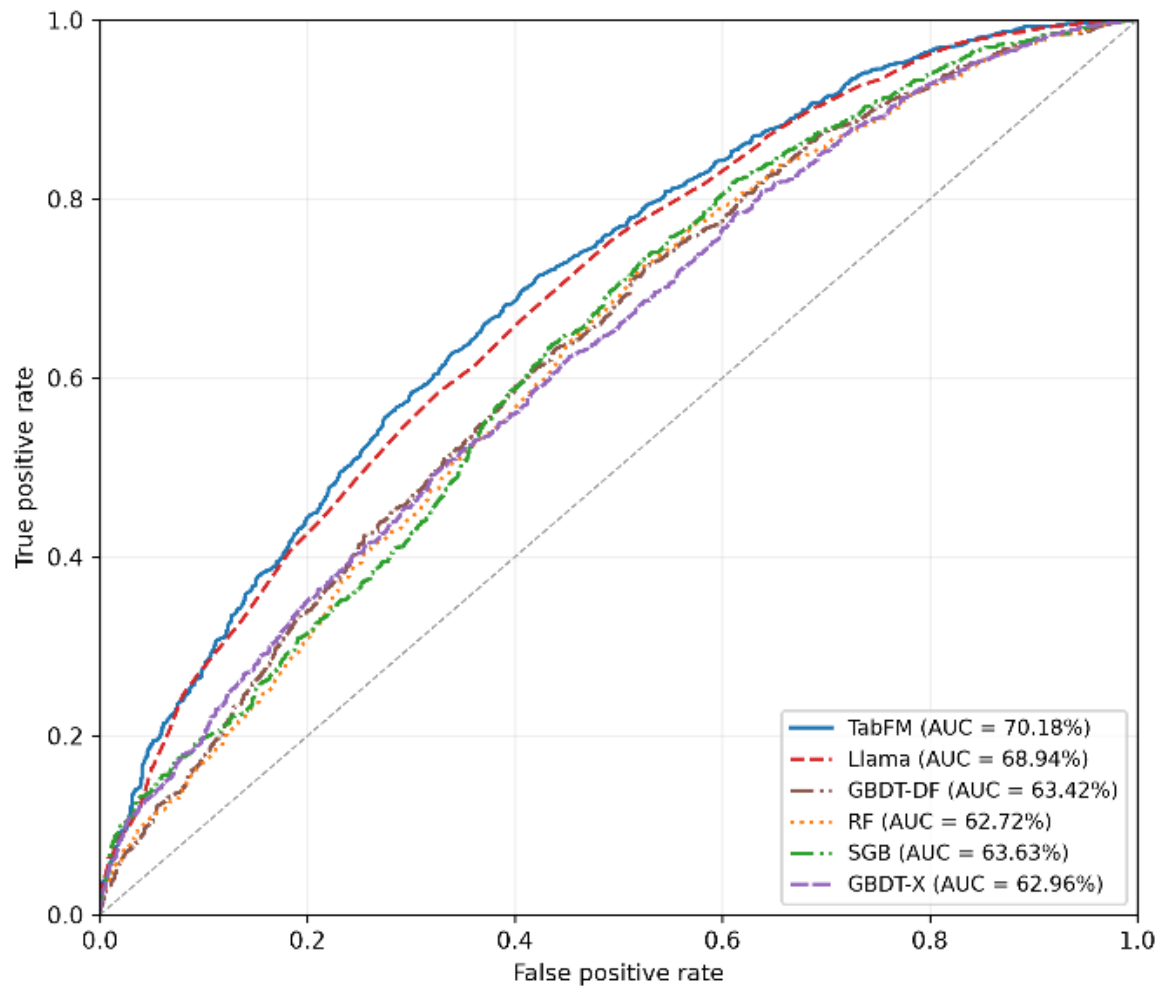

*Panel C: 2024*

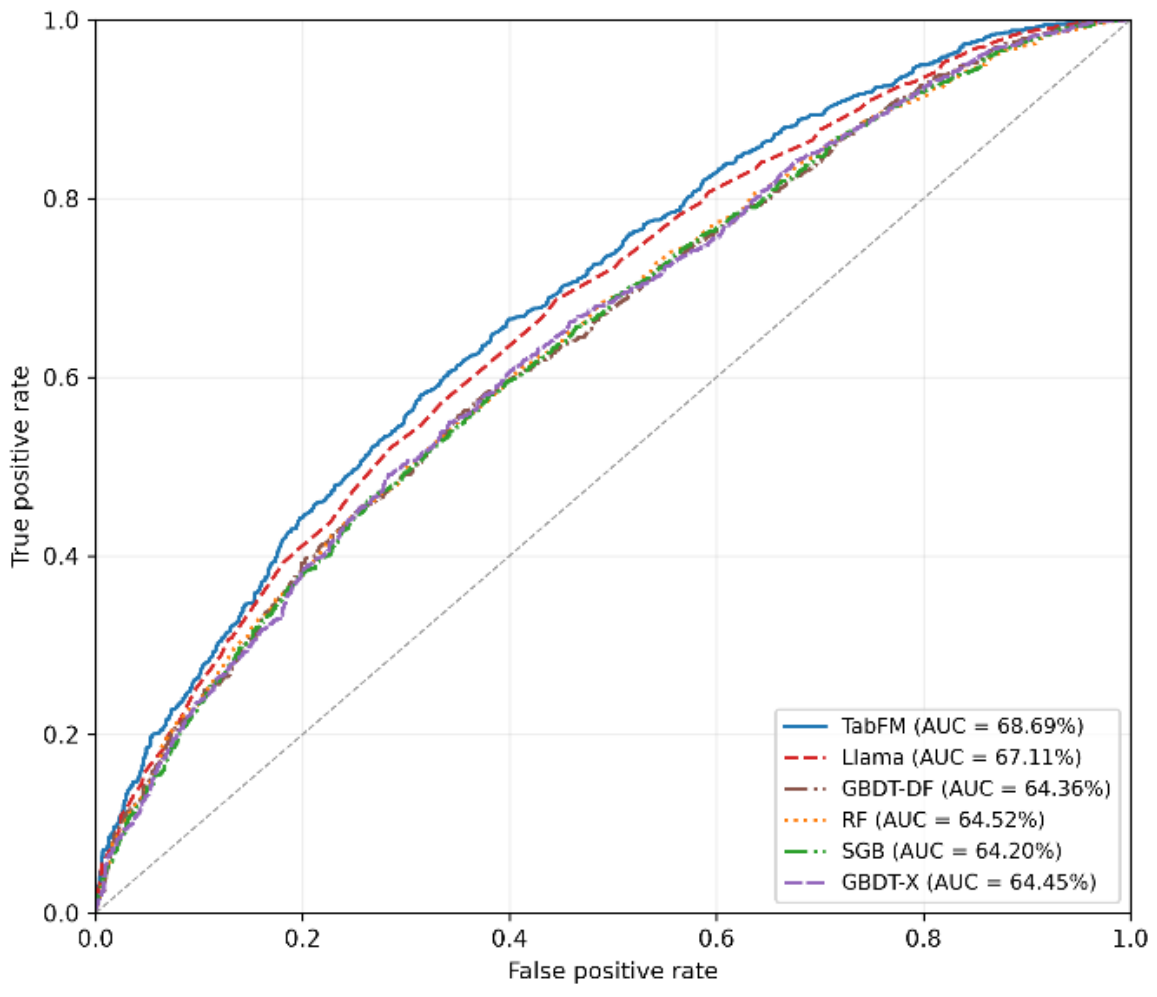

**Figure 2. AUCs Under Numerical Perturbations**

This figure compares AUCs before and after numerical perturbations to the test predictors. Each perturbation randomly replaces up to one, three, or five digits at the end of each test predictor's recorded value. For TabFM, RF, SGB, and GBDT-X, the bars show AUCs averaged across ten random draws, using the same test observations in every condition. These four models receive identical perturbed test values, while their training data, validation data, and in-context training samples are unchanged. Llama's results are taken from DF (Table 7, Panel A). For reference, the figure also reports Levy's (2026, Tables 6 and 7, Panel A) results for GPT-4 and gradient-boosted decision tree (GBDT-L), which are based on a different sample than ours. See Section 3.5 for discussion.

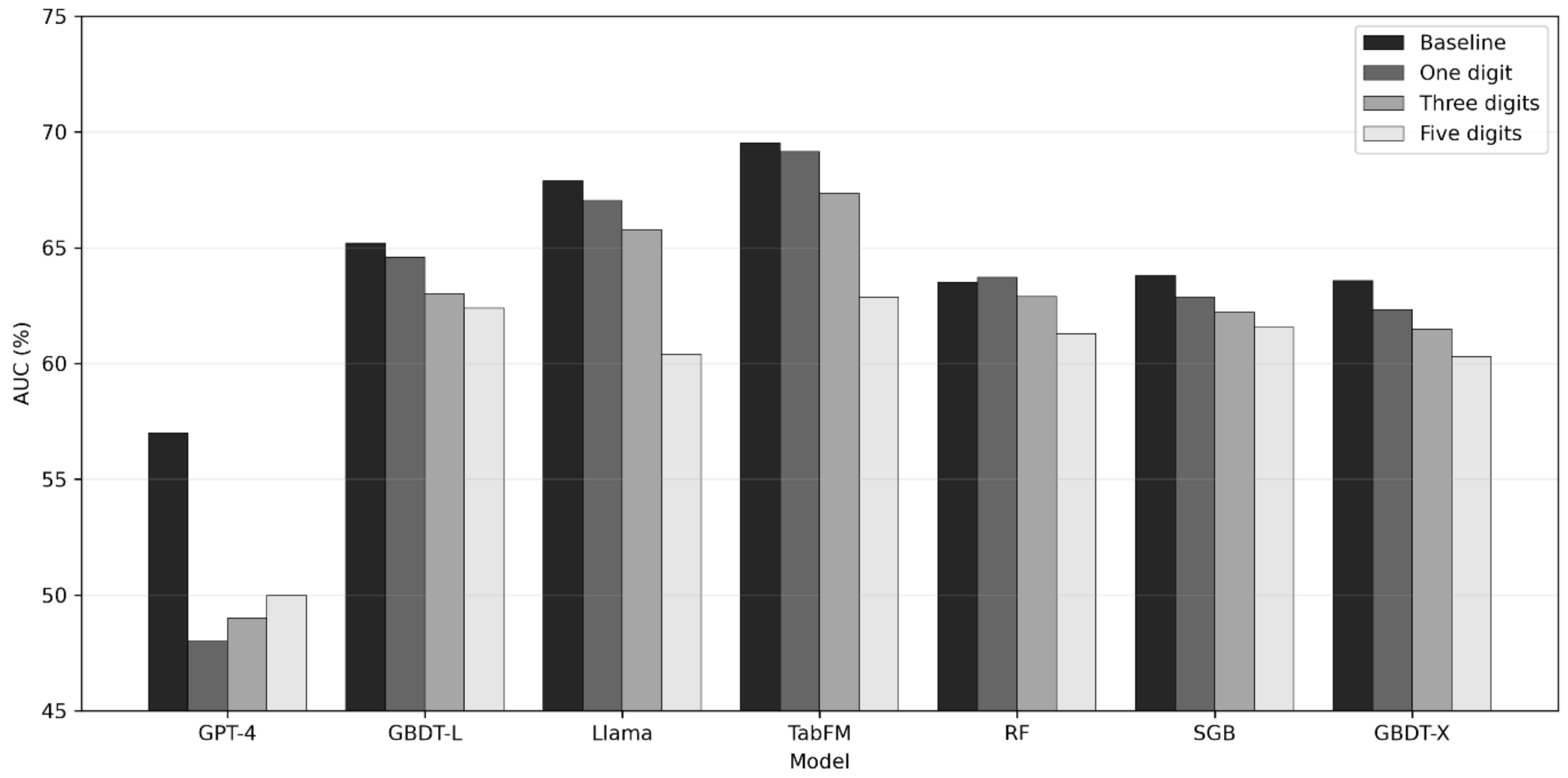

**Figure 3. Annual AUCs**

This figure compares annual AUCs for TabFM, RF, SGB, and GBDT-X for financial statement years 2011–2024. Table 3 reports pooled performance and pairwise comparisons. See Section 4.1 for discussion.

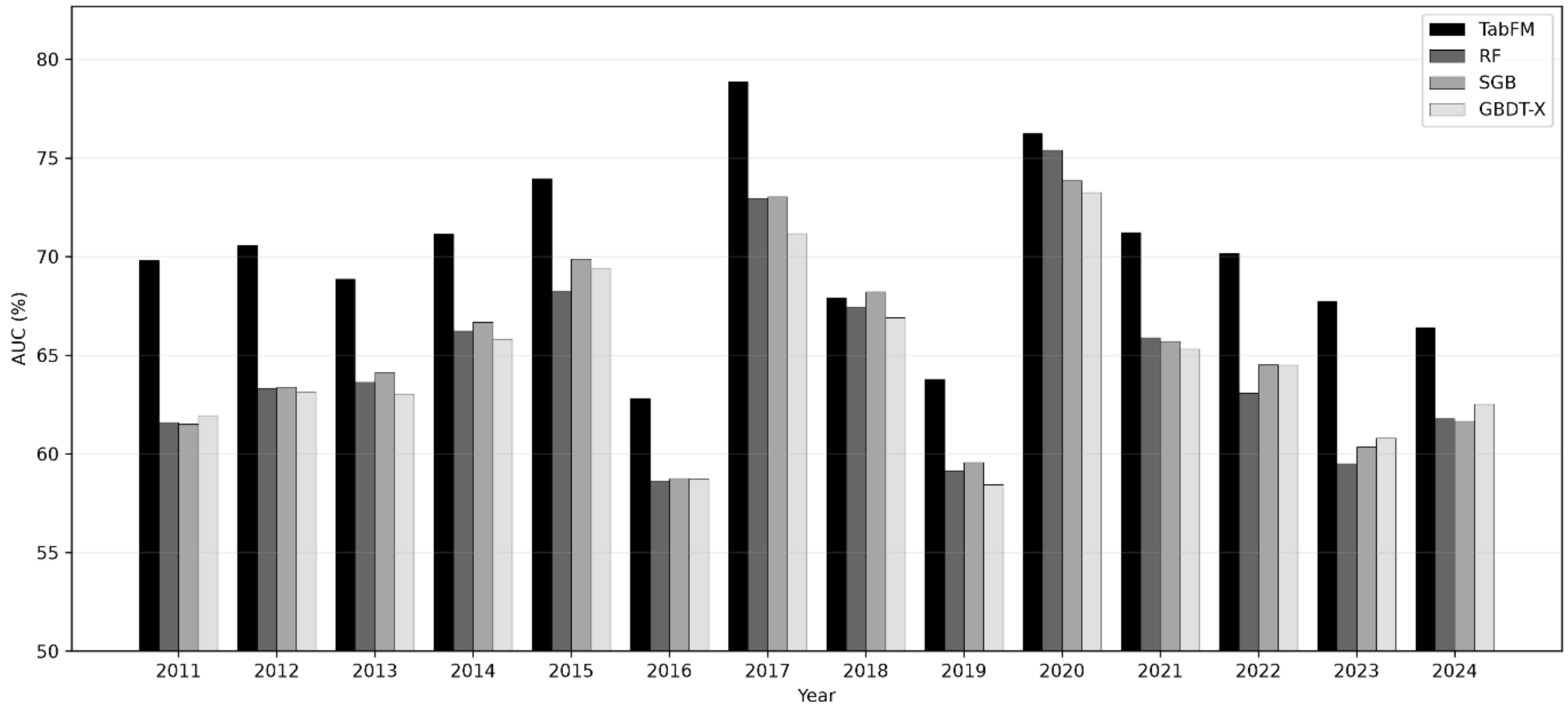

## Table 1. Annual Distribution of Outcomes

This table reports the annual distribution of earnings increases for the sample supplied by DF. The sample contains 62,796 firm-year observations, covering financial statement years 2005–2024. For each financial statement year *t*, the table reports the number and percentage of observations whose net income increases in year *t*+1 and of those whose net income does not increase. DF (Table 2, Panel B) report the same distribution indexed by year *t*+1.

| Financial Statement Year | Increases | Non-increases | Total | % Increases | % Non-increases |
|---|---|---|---|---|---|
| 2005 | 2,101 | 1,276 | 3,377 | 62.21 | 37.79 |
| 2006 | 1,797 | 1,542 | 3,339 | 53.82 | 46.18 |
| 2007 | 1,093 | 1,962 | 3,055 | 35.78 | 64.22 |
| 2008 | 1,552 | 1,562 | 3,114 | 49.84 | 50.16 |
| 2009 | 2,137 | 995 | 3,132 | 68.23 | 31.77 |
| 2010 | 1,833 | 1,155 | 2,988 | 61.35 | 38.65 |
| 2011 | 1,661 | 1,343 | 3,004 | 55.29 | 44.71 |
| 2012 | 1,780 | 1,295 | 3,075 | 57.89 | 42.11 |
| 2013 | 1,776 | 1,296 | 3,072 | 57.81 | 42.19 |
| 2014 | 1,543 | 1,511 | 3,054 | 50.52 | 49.48 |
| 2015 | 1,775 | 1,319 | 3,094 | 57.37 | 42.63 |
| 2016 | 1,833 | 1,363 | 3,196 | 57.35 | 42.65 |
| 2017 | 1,904 | 1,214 | 3,118 | 61.06 | 38.94 |
| 2018 | 1,681 | 1,401 | 3,082 | 54.54 | 45.46 |
| 2019 | 1,328 | 1,816 | 3,144 | 42.24 | 57.76 |
| 2020 | 2,305 | 843 | 3,148 | 73.22 | 26.78 |
| 2021 | 1,505 | 1,564 | 3,069 | 49.04 | 50.96 |
| 2022 | 1,543 | 1,623 | 3,166 | 48.74 | 51.26 |
| 2023 | 1,794 | 1,576 | 3,370 | 53.23 | 46.77 |
| 2024 | 1,867 | 1,332 | 3,199 | 58.36 | 41.64 |
| **Total** | **34,808** | **27,988** | **62,796** | **55.43** | **44.57** |

**Table 2. Predictive Performance and Computational Requirements**

This table compares the predictive performance of TabFM, Llama, GBDT-DF, RF, SGB, and GBDT-X in forecasting next-year earnings increases and reports computational requirements for all six models. The comparison uses the full set of predictors and test observations based on 2023 and 2024 financial statements. The Llama and GBDT-DF values in Panels A to C are computed from the observation-level predictions provided by DF. Panel A reports pooled AUC, overall classification accuracy, increase accuracy, and non-increase accuracy. An observation is classified as an increase when its predicted probability exceeds 0.5, as in DF. Panel B reports pairwise differences in pooled AUC, computed as the row model minus the column model. The *p*-values in parentheses are from the paired bootstrap described in Section 3.3. Panel C reports AUCs separately for the two test years. Panel D reports running time by task, peak memory use, and transformer-based models' FLOP. Times are in minutes and cover both test years. TabFM runs on one GPU, an NVIDIA GeForce RTX 4080 SUPER with memory expanded to about 32 GiB. SGB and GBDT-X run on the same GPU with four CPU threads. RF runs on the CPU, fitting up to eight trees in parallel. The Llama and GBDT-DF figures are approximate values provided by DF. DF fine-tune and apply Llama on one NVIDIA H200 GPU, and they run the GBDT-DF search on a server with 128 CPU cores. The abbreviation n/a denotes not applicable. Computational measurements follow Appendix A.3.

*Panel A: Overall predictive performance*

| Metric | TabFM | Llama | GBDT-DF | RF | SGB | GBDT-X |
|---|---|---|---|---|---|---|
| Pooled AUC (%) | 69.53 | 67.88 | 62.75 | 63.51 | 63.78 | 63.57 |
| Overall classification accuracy (%) | 64.32 | 62.43 | 60.19 | 61.06 | 60.97 | 59.57 |
| Increase accuracy (%) | 73.18 | 63.75 | 69.16 | 73.37 | 71.48 | 67.99 |
| Non-increase accuracy (%) | 53.16 | 60.76 | 48.90 | 45.56 | 47.73 | 48.97 |

*Panel B: Pairwise differences in pooled AUC (with p-values in parentheses)*

| Model | Llama | GBDT-DF | RF | SGB | GBDT-X |
|---|---|---|---|---|---|
| TabFM | 1.66 | 6.78 | 6.02 | 5.75 | 5.96 |
| | (0.001) | (<0.001) | (<0.001) | (<0.001) | (<0.001) |
| Llama | | 5.13 | 4.36 | 4.10 | 4.31 |
| | | (<0.001) | (<0.001) | (<0.001) | (<0.001) |
| GBDT-DF | | | -0.76 | -1.03 | -0.82 |
| | | | (0.119) | (0.025) | (0.110) |
| RF | | | | -0.26 | -0.05 |
| | | | | (0.534) | (0.902) |
| SGB | | | | | 0.21 |
| | | | | | (0.653) |

*Panel C: Annual predictive performance*

| Metric | TabFM | Llama | GBDT-DF | RF | SGB | GBDT-X |
|---|---|---|---|---|---|---|
| AUC (%) in 2023 | 70.18 | 68.94 | 63.42 | 62.72 | 63.63 | 62.96 |
| AUC (%) in 2024 | 68.69 | 67.11 | 64.36 | 64.52 | 64.20 | 64.45 |

*Panel D: Computational requirements*

| Model | Training (min) | Predicting (min) | Total (min) | Peak RAM (GiB) | Peak GPU memory (GiB) | FLOP |
|---|---|---|---|---|---|---|
| TabFM | n/a | 11.83 | 11.83 | 13.69 | 24.27 | $1.817 \times 10^{16}$ |
| Llama (from DF, estimated) | ≈6,480 | ≈6 | ≈6,486 | not reported | 134 | $1.2 \times 10^{20}$ |
| GBDT-DF (from DF, estimated) | ≈7,200 | not reported | ≈7,200 | not reported | n/a | n/a |
| RF | 130.07 | 0.09 | 130.16 | 7.18 | n/a | n/a |
| SGB | 7.51 | 0.01 | 7.52 | 0.63 | 0.37 | n/a |
| GBDT-X | 139.75 | 0.10 | 139.84 | 4.15 | 2.28 | n/a |

## Table 3. Predictive Performance Across Years

This table reports pooled forecasting performance and its variation across years for TabFM, RF, SGB, and GBDT-X for financial statement years 2011–2024. For each test year, RF, SGB, and GBDT-X use five estimation years and one validation year, and TabFM uses the combined six years as its in-context training sample. Figure 3 reports the annual AUCs. Panel A reports pooled AUC, standard deviations of annual AUCs, and summaries of TabFM's annual differences from RF, SGB, and GBDT-X, including the number of years with a positive difference. Panel B reports pairwise differences in pooled AUC, computed as the row model minus the column model, with *p*-values in parentheses from the paired bootstrap described in Section 3.3.

*Panel A: Pooled AUC and variation across years*

| Metric | TabFM | RF | SGB | GBDT-X |
|---|---|---|---|---|
| Pooled AUC (%) | 69.52 | 64.57 | 64.82 | 64.68 |
| Std. dev. of annual AUC (pp) | 4.39 | 5.00 | 4.78 | 4.39 |
| | | | | |
| TabFM's average annual advantage (pp) | | 5.18 | 4.88 | 5.33 |
| Std. dev. of TabFM's annual advantage (pp) | - | 2.32 | 2.15 | 1.92 |
| Largest annual advantage (pp) | - | 8.25 | 8.30 | 7.86 |
| Smallest annual advantage (pp) | - | 0.48 | -0.29 | 1.00 |
| Years TabFM's AUC is higher | - | 14 of 14 | 13 of 14 | 14 of 14 |

*Panel B: Pairwise differences in pooled AUC (with p-values in parentheses)*

| Model | RF | SGB | GBDT-X |
|---|---|---|---|
| TabFM | 4.96 | 4.70 | 4.84 |
| | (<0.001) | (<0.001) | (<0.001) |
| RF | | -0.26 | -0.11 |
| | | (0.099) | (0.523) |
| SGB | | | 0.15 |
| | | | (0.331) |

## Table 4. Sensitivity to Omitting Features

This table evaluates forecasting performance for TabFM, RF, SGB, and GBDT-X after removing some of the original 130 predictors. Performance is evaluated on the same test observations as in Table 2. The table reports the baseline using all 130 predictors and five omission conditions. The first removes the current-year values of five income measures: net income; income before extraordinary items and noncontrolling interest; income before extraordinary items; income before extraordinary items for common stocks; and income before extraordinary items adjusted for common stock equivalents. The second also removes their values for $t-1$ and $t-2$. The third removes the 20 predictors that DF rank as most important (DF, Table 5). The last two remove the $t-2$ income statement and $t-1$ balance sheet, and the entire balance sheet together with the $t-2$ income statement. RF, SGB, and GBDT-X are re-estimated with configurations selected again in each condition. ΔAUC is the omission condition minus the baseline, computed from unrounded AUCs. The *p*-values are from the paired bootstrap described in Section 3.3.

| | **TabFM** | **RF** | **SGB** | **GBDT-X** |
|---|---|---|---|---|
| **Baseline (130 predictors)** | | | | |
| AUC (%) | 69.53 | 63.51 | 63.78 | 63.57 |
| **Five measures omitted (125 predictors)** | | | | |
| AUC (%) | 69.48 | 63.32 | 63.50 | 63.51 |
| Δ AUC (pp) | -0.06 | -0.19 | -0.28 | -0.06 |
| *p*-value | (0.747) | (0.301) | (0.289) | (0.888) |
| **Five measures omitted, *t* to *t*−2 (115 predictors)** | | | | |
| AUC (%) | 68.45 | 62.60 | 63.07 | 63.25 |
| Δ AUC (pp) | -1.09 | -0.91 | -0.71 | -0.32 |
| *p*-value | (<0.001) | (<0.001) | (0.022) | (0.481) |
| **DF's top 20 omitted (110 predictors)** | | | | |
| AUC (%) | 68.22 | 61.46 | 61.28 | 61.62 |
| Δ AUC (pp) | -1.31 | -2.06 | -2.50 | -1.95 |
| *p*-value | (<0.001) | (<0.001) | (<0.001) | (<0.001) |
| ***t*−1 balance sheet and *t*−2 income statement omitted (78 predictors)** | | | | |
| AUC (%) | 69.97 | 64.46 | 63.98 | 63.94 |
| Δ AUC (pp) | +0.44 | +0.95 | +0.20 | +0.37 |
| *p*-value | (0.199) | (0.002) | (0.581) | (0.465) |
| **Balance sheet and *t*−2 income statement omitted (52 predictors)** | | | | |
| AUC (%) | 69.59 | 63.11 | 62.43 | 62.06 |
| Δ AUC (pp) | +0.05 | -0.40 | -1.35 | -1.51 |
| *p*-value | (0.899) | (0.300) | (0.024) | (0.010) |

**Table 5. Performance With Smaller Training Samples**

This table examines forecasting performance for TabFM, RF, SGB, and GBDT-X when fewer training observations are available. Panels A and B reduce the share of training observations in the principal sample. The reduced samples contain 80% to 5% of the observations in each year, drawn without replacement, and are evaluated on the same test observations as in Table 2. The tree-based models retain their full validation year. Panels C and D reduce the number of firms. For each of the 2023 and 2024 test years, we randomly select 500, 400, 300, 200, or 100 firms, and retain their available observations over the preceding six years, with five estimation years and one validation year for the tree-based models. Every condition is evaluated on the test-year observations of the same 500 firms in each test year, 1,000 observations in total. In both tests, the samples are nested, and the tree-based models select their configurations again in each condition. Values are averages across ten random draws, except the 100% values in Panels A and B, which are from Table 2. TabFM uses the estimation and validation years combined as its in-context training sample, and Training # reports its size for each test year. ΔAUC is the AUC in the largest sample minus the AUC in the smallest, in percentage points. Skill lost is ΔAUC divided by the largest-sample AUC minus 50, the AUC of uninformative forecasts. The change rows compare the smallest sample with the largest, in percent. Computational measurements follow Appendix A.3.

*Panel A: AUC (%) by share of training sample*

| | Share of training sample | | | | | | | | |
|---|---|---|---|---|---|---|---|---|---|
| **Model** | **100%** | **80%** | **60%** | **40%** | **20%** | **10%** | **5%** | **ΔAUC, 100% to 5% (pp)** | **Skill lost (%)** |
| **Training #:** | | | | | | | | | |
| **2023** | 56,227 | 44,981 | 33,734 | 22,493 | 11,246 | 5,623 | 2,812 | | |
| **2024** | 56,220 | 44,975 | 33,730 | 22,490 | 11,245 | 5,622 | 2,811 | | |
| TabFM | 69.53 | 69.20 | 68.90 | 68.36 | 67.64 | 66.30 | 64.47 | 5.06 | 25.90 |
| RF | 63.51 | 63.38 | 63.12 | 62.77 | 61.86 | 61.02 | 59.50 | 4.01 | 29.68 |
| SGB | 63.78 | 63.58 | 63.27 | 62.79 | 61.98 | 61.11 | 60.24 | 3.54 | 25.70 |
| GBDT-X | 63.57 | 63.39 | 63.00 | 62.54 | 61.00 | 60.01 | 59.62 | 3.95 | 29.09 |

*Panel B: Computational requirements by share of training sample*

| | AUC (%) | Training (min) | Predicting (min) | Total (min) | Peak RAM (GiB) | Peak GPU memory (GiB) |
|---|---|---|---|---|---|---|
| **TabFM** | | | | | | |
| 100% | 69.53 | n/a | 11.83 | 11.83 | 13.69 | 24.27 |
| 80% | 69.20 | n/a | 7.85 | 7.85 | 13.51 | 20.68 |
| 60% | 68.90 | n/a | 5.69 | 5.69 | 13.48 | 17.38 |
| 40% | 68.36 | n/a | 3.85 | 3.85 | 13.49 | 13.82 |
| 20% | 67.64 | n/a | 2.31 | 2.31 | 13.50 | 9.35 |
| 10% | 66.30 | n/a | 1.59 | 1.59 | 13.60 | 8.26 |
| 5% | 64.47 | n/a | 1.28 | 1.28 | 13.44 | 7.85 |
| **Change, 100% to 5%** | **-7.3%** | **-** | **-89.1%** | **-89.1%** | **-1.8%** | **-67.7%** |
| **RF** | | | | | | |
| 100% | 63.51 | 130.07 | 0.09 | 130.16 | 7.18 | n/a |
| 80% | 63.38 | 98.86 | 0.10 | 98.96 | 7.08 | n/a |
| 60% | 63.12 | 72.40 | 0.08 | 72.49 | 5.71 | n/a |
| 40% | 62.77 | 46.03 | 0.06 | 46.10 | 4.33 | n/a |
| 20% | 61.86 | 22.21 | 0.03 | 22.24 | 3.01 | n/a |
| 10% | 61.02 | 12.83 | 0.03 | 12.85 | 2.46 | n/a |
| 5% | 59.50 | 7.25 | 0.02 | 7.27 | 1.99 | n/a |
| **Change, 100% to 5%** | **-6.3%** | **-94.4%** | **-79.4%** | **-94.4%** | **-72.3%** | **-** |
| **SGB** | | | | | | |
| 100% | 63.78 | 7.51 | 0.01 | 7.52 | 0.63 | 0.37 |
| 80% | 63.58 | 6.80 | 0.01 | 6.81 | 0.66 | 0.36 |
| 60% | 63.27 | 6.13 | 0.01 | 6.14 | 0.67 | 0.36 |
| 40% | 62.79 | 5.27 | 0.01 | 5.28 | 0.60 | 0.36 |
| 20% | 61.98 | 4.50 | 0.01 | 4.51 | 0.63 | 0.36 |
| 10% | 61.11 | 4.24 | 0.01 | 4.25 | 0.65 | 0.36 |
| 5% | 60.24 | 3.90 | 0.01 | 3.91 | 0.55 | 0.35 |
| **Change, 100% to 5%** | **-5.6%** | **-48.1%** | **-13.2%** | **-48.0%** | **-12.7%** | **-4.3%** |
| **GBDT-X** | | | | | | |
| 100% | 63.57 | 139.75 | 0.10 | 139.84 | 4.15 | 2.28 |
| 80% | 63.39 | 128.07 | 0.09 | 128.16 | 3.79 | 2.21 |
| 60% | 63.00 | 121.22 | 0.09 | 121.32 | 3.40 | 2.22 |
| 40% | 62.54 | 108.96 | 0.09 | 109.05 | 3.13 | 2.17 |
| 20% | 61.00 | 92.80 | 0.09 | 92.89 | 2.54 | 2.08 |
| 10% | 60.01 | 80.05 | 0.09 | 80.14 | 2.29 | 1.98 |
| 5% | 59.62 | 70.27 | 0.09 | 70.35 | 1.89 | 1.89 |
| **Change, 100% to 5%** | **-6.2%** | **-49.7%** | **-11.4%** | **-49.7%** | **-54.5%** | **-16.9%** |

*Panel C: AUC (%) by number of firms*

| Model | # of firms 500 | 400 | 300 | 200 | 100 | ΔAUC, 500-100 (pp) | Skill lost (%) |
|---|---|---|---|---|---|---|---|
| **Average Training #:** | | | | | | | |
| **2023** | 2,227 | 1,781 | 1,336 | 886 | 447 | | |
| **2024** | 2,323 | 1,865 | 1,398 | 932 | 466 | | |
| TabFM | 62.92 | 62.42 | 61.02 | 60.88 | 59.51 | 3.41 | 26.39 |
| RF | 57.33 | 57.19 | 56.85 | 56.33 | 55.94 | 1.39 | 18.97 |
| SGB | 57.98 | 57.85 | 57.22 | 56.77 | 56.15 | 1.83 | 22.95 |
| GBDT-X | 56.69 | 57.61 | 56.66 | 55.13 | 54.69 | 2.00 | 29.83 |

*Panel D: Computational requirements by number of firms*

| | AUC (%) | Training (min) | Predicting (min) | Total (min) | Peak RAM (GiB) | Peak GPU memory (GiB) |
|---|---|---|---|---|---|---|
| **TabFM** | | | | | | |
| 500 | 62.92 | n/a | 0.84 | 0.84 | 13.30 | 7.85 |
| 400 | 62.42 | n/a | 0.95 | 0.95 | 13.42 | 7.21 |
| 300 | 61.02 | n/a | 0.83 | 0.83 | 13.45 | 6.50 |
| 200 | 60.88 | n/a | 0.69 | 0.69 | 13.32 | 5.78 |
| 100 | 59.51 | n/a | 0.65 | 0.65 | 13.40 | 5.07 |
| **Change, 500 to 100** | **-5.4%** | **-** | **-22.3%** | **-22.3%** | **+0.7%** | **-35.5%** |
| **RF** | | | | | | |
| 500 | 57.33 | 5.07 | 0.02 | 5.09 | 1.87 | n/a |
| 400 | 57.19 | 4.44 | 0.09 | 4.53 | 1.83 | n/a |
| 300 | 56.85 | 3.68 | 0.02 | 3.70 | 1.78 | n/a |
| 200 | 56.33 | 2.87 | 0.01 | 2.88 | 1.73 | n/a |
| 100 | 55.94 | 2.17 | 0.02 | 2.18 | 1.70 | n/a |
| **Change, 500 to 100** | **-2.4%** | **-57.3%** | **-10.6%** | **-57.1%** | **-9.3%** | **-** |
| **SGB** | | | | | | |
| 500 | 57.98 | 3.64 | 0.01 | 3.65 | 0.58 | 0.36 |
| 400 | 57.85 | 3.62 | 0.01 | 3.63 | 0.58 | 0.36 |
| 300 | 57.22 | 3.46 | 0.03 | 3.50 | 0.58 | 0.36 |
| 200 | 56.77 | 3.38 | 0.01 | 3.39 | 0.57 | 0.36 |
| 100 | 56.15 | 3.13 | 0.01 | 3.14 | 0.57 | 0.36 |
| **Change, 500 to 100** | **-3.2%** | **-14.0%** | **+3.2%** | **-14.0%** | **-1.4%** | **-0.6%** |
| **GBDT-X** | | | | | | |
| 500 | 56.69 | 63.62 | 0.03 | 63.65 | 1.17 | 1.71 |
| 400 | 57.61 | 60.19 | 0.03 | 60.22 | 1.10 | 1.69 |
| 300 | 56.66 | 56.05 | 0.03 | 56.07 | 0.98 | 1.43 |
| 200 | 55.13 | 50.83 | 0.02 | 50.86 | 0.90 | 1.24 |
| 100 | 54.69 | 42.66 | 0.03 | 42.69 | 0.72 | 0.99 |
| **Change, 500 to 100** | **-3.5%** | **-32.9%** | **-4.1%** | **-32.9%** | **-38.1%** | **-42.1%** |

## Table 6. Evaluation of Alternative Tabular Foundation Models

This table examines TabICLv2, TabPFN-3.5 and alternative TabFM configurations and versions. The forecasting task is the same as in Table 2. TabFM-1.0.0 (8 members) is the main specification used in Tables 2 to 5. It uses eight ensemble members. TabFM-1.0.0 (32 members) uses 32 members. TabFM+ is an enhanced configuration, which adds engineered predictors to half of the 32 members, weights the members using a validation portion of the in-context training sample, and calibrates the probabilities. TabFM-1.1.0 rows use a later release of the pretrained model under the same three configurations (Appendix A.1.1). Panel A reports pooled AUC, running time, peak memory use, and FLOP. The TabFM-1.0.0 (8 members) results reproduce Table 2. Panel B reports pairwise differences in pooled AUC, computed as the row model minus the column model, with *p*-values in parentheses from the paired bootstrap described in Section 3.3. Computational measurements follow Appendix A.3.

*Panel A: Pooled AUC (%) and computational requirements*

| Model | Pooled AUC (%) | Predicting (min) | Peak RAM (GiB) | Peak GPU memory (GiB) | FLOP |
|---|---|---|---|---|---|
| TabFM-1.0.0 (8 members) | 69.53 | 11.83 | 13.69 | 24.27 | $1.817 \times 10^{16}$ |
| TabICLv2 | 67.57 | 1.17 | 2.52 | 30.46 | $1.657 \times 10^{15}$ |
| TabPFN-3.5 | 68.55 | 2.57 | 3.96 | 3.79 | $6.087 \times 10^{15}$ |
| TabFM-1.0.0 (32 members) | 69.48 | 41.18 | 13.73 | 24.27 | $7.269 \times 10^{16}$ |
| TabFM+-1.0.0 (32 members) | 69.13 | 82.72 | 13.73 | 27.81 | $1.441 \times 10^{17}$ |
| TabFM-1.1.0 (8 members) | 68.36 | 5.59 | 3.76 | 10.42 | $7.530 \times 10^{15}$ |
| TabFM-1.1.0 (32 members) | 68.24 | 19.52 | 4.99 | 10.42 | $3.012 \times 10^{16}$ |
| TabFM+-1.1.0 (32 members) | 68.09 | 38.34 | 5.64 | 11.98 | $5.891 \times 10^{16}$ |

*Panel B: Pairwise differences in pooled AUC (p-values in parentheses)*

| Model | TabFM-1.0.0 (8 members) | Llama | GBDT-DF | RF | SGB | GBDT-X |
|---|---|---|---|---|---|---|
| TabFM-1.0.0 (8 members) | | 1.66 | 6.78 | 6.02 | 5.75 | 5.96 |
| | | (0.001) | (<0.001) | (<0.001) | (<0.001) | (<0.001) |
| TabICLv2 | -1.96 | -0.31 | 4.82 | 4.06 | 3.79 | 4.00 |
| | (<0.001) | (0.554) | (<0.001) | (<0.001) | (<0.001) | (<0.001) |
| TabPFN-3.5 | -0.98 | 0.67 | 5.80 | 5.04 | 4.77 | 4.98 |
| | (<0.001) | (0.177) | (<0.001) | (<0.001) | (<0.001) | (<0.001) |
| TabFM-1.0.0 (32 members) | -0.05 | 1.60 | 6.73 | 5.96 | 5.70 | 5.91 |
| | (0.442) | (0.002) | (<0.001) | (<0.001) | (<0.001) | (<0.001) |
| TabFM+-1.0.0 (32 members) | −0.40 | 1.26 | 6.38 | 5.62 | 5.36 | 5.57 |
| | (0.004) | (0.013) | (<0.001) | (<0.001) | (<0.001) | (<0.001) |
| TabFM-1.1.0 (8 members) | −1.17 | 0.48 | 5.61 | 4.85 | 4.58 | 4.79 |
| | (<0.001) | (0.345) | (<0.001) | (<0.001) | (<0.001) | (<0.001) |
| TabFM-1.1.0 (32 members) | −1.29 | 0.36 | 5.49 | 4.73 | 4.46 | 4.67 |
| | (<0.001) | (0.476) | (<0.001) | (<0.001) | (<0.001) | (<0.001) |
| TabFM+-1.1.0 (32 members) | −1.44 | 0.22 | 5.34 | 4.58 | 4.32 | 4.53 |
| | (<0.001) | (0.673) | (<0.001) | (<0.001) | (<0.001) | (<0.001) |